\documentclass{article}

\PassOptionsToPackage{square, numbers, compress}{natbib}

\usepackage[final]{neurips_2021}

\usepackage[utf8]{inputenc} %
\usepackage[T1]{fontenc}    %
\usepackage[hidelinks]{hyperref}       %
\usepackage{url}            %
\usepackage{booktabs}       %
\usepackage{amsfonts}       %
\usepackage{nicefrac}       %
\usepackage{microtype}      %
\usepackage{xcolor}         %

\usepackage{times}
\usepackage{epsfig}
\usepackage{graphicx}
\usepackage{amsmath}
\usepackage{amssymb}
\usepackage{amssymb}
\usepackage{subfiles}
\usepackage{multirow}
\usepackage[toc,page]{appendix}
\usepackage{floatrow}
\usepackage[algo2e,ruled,linesnumbered,lined]{algorithm2e}
\usepackage{amsmath}
\usepackage{mathtools}
\usepackage{wrapfig}
\usepackage{bbm}
\usepackage{color, colortbl}
\definecolor{Gray}{gray}{0.95}
\usepackage{subcaption}
\DeclareMathOperator*{\argmin}{arg\,min}

\newcommand\blfootnote[1]{%
  \begingroup
  \renewcommand\thefootnote{}\footnote{#1}%
  \addtocounter{footnote}{-1}%
  \endgroup
}

\title{Continual Learning via Local Module Composition}

\author{%
    Oleksiy Ostapenko$^{12}$ \quad Pau Rodr\'iguez$^3$ \quad Massimo Caccia$^{123}$ \quad Laurent Charlin$^{145}$ \\
    $^1$Mila - Quebec AI Institute, $^2$Université de Montréal, $^3$ServiceNow,  $^4$HEC Montréal, \\ $^5$Canada CIFAR AI Chair \\
}
\begin{document}

\maketitle

\begin{abstract} 
Modularity is a compelling solution to continual learning (CL), the problem of modeling sequences of related tasks. Learning and then composing modules to solve different tasks provides an abstraction to address the principal challenges of CL including catastrophic forgetting, backward and forward transfer across tasks, and sub-linear model growth. We introduce local module composition (LMC), an approach to modular CL where each module is provided a local structural component that estimates a module's relevance to the input. Dynamic module composition is performed layer-wise based on local relevance scores. We demonstrate that agnosticity to task identities (IDs) arises from (local) structural learning that is module-specific as opposed to the task- and/or model-specific as in previous works, making LMC applicable to more CL settings compared to previous works. In addition, LMC also tracks statistics about the input distribution and adds new modules when outlier samples are detected. In the first set of experiments, LMC performs favorably compared to existing methods on the recent Continual Transfer-learning Benchmark without requiring task identities. In another study, we show that the locality of structural learning allows LMC to interpolate to related but unseen tasks (OOD), as well as to compose modular networks trained independently on different task sequences into a third modular network without any fine-tuning. Finally, in search for limitations of LMC we study it on more challenging sequences of 30 and 100 tasks, demonstrating that local module selection becomes much more challenging in presence of a large number of candidate modules. In this setting best performing LMC spawns much fewer modules compared to an oracle based baseline, however it reaches a lower overall accuracy. The codebase is available under \url{https://github.com/oleksost/LMC}. 
\blfootnote{corresponding author: oleksiy.ostapenko@t-online.de}
\end{abstract}

\section{Introduction}
The goal of continual learning (CL) is to learn efficiently from a non-stationary stream of tasks without (catastrophically) forgetting previous tasks~\cite{mccloskey1989catastrophic}. CL is often modeled as a trade-off between knowledge retention (stability) and knowledge expansion (plasticity) \cite{french1997pseudo,mermillod2013stability}. Parameter sharing can provide control over this trade-off. For example, learning a single model shared across tasks results in better knowledge transfer and faster learning at the expense of forgetting~\citep{kirkpatrick2017overcoming,li2017learning}. Conversely, learning a separate model per task eliminates forgetting but minimizes transfer and data efficiency~\citep{aljundi2017expert,jerfel2019reconciling}.  

Modular learning aims at balancing transfer and forgetting by learning a set of specialized modules that can be recomposed to solve (new) tasks while only updating a subset of relevant modules or adding new modules~\citep{andreas2016neural,kirsch2018modular,goyal2019recurrent}. In principle, a modular learner capable of composing modules in meaningful structures can provide additional benefits including \textbf{(i)} computational gains due to only executing modules that are relevant to a task~\cite{kirsch2018modular,amer2019review}; \textbf{(ii)} memory gains due to instantiating a sub-linear number of modules w.r.t. the number of tasks; \textbf{(iii)} systematic~\cite{bahdanau2018systematic} and out-of-distribution (OOD) generalization~\cite{corona2020modularity} through knowledge recombination; %
and \textbf{(iv)} biological plausibility~\cite{sternberg2011modular, sporns2016modular,whittington2017approximation}.

Designing modular methods for CL comes with two main challenges. The first is \emph{how and when to add new modules} to ensure sufficient plasticity to learn new tasks. Existing modular methods use greedy search variants, expanding the model when it improves validation performance~\citep{veniat2020efficient,mendez2020lifelong}. The second challenge is \emph{how to compose} that is, retrieve task-specific structural knowledge given a new task (previously seen or not).

Existing methods rely on a task's identifier (ID) to retrieve task-specific structural knowledge, which comes either in the form of an optimal module layout~\cite{veniat2020efficient} or as a model- and task-specific controller network that generates modular layouts~\cite{mendez2020lifelong}. Unfortunately, in many realistic CL scenarios task identities are unavailable at test time~\cite{Farquhar18,he2019task,caccia2020online}.
Lifting this limitation is challenging since standard mechanisms for task inference, for example, leveraging a task-inference model, could be subject to forgetting themselves.

To address both challenges, we equip each module with a \emph{local structural component} that predicts a score indicating how relevant the module is for a given input. In-distribution inputs result in high scores, while out-of-distribution inputs result in low scores. In other words, modules self-determine their relevance given an input. 

This local component is used for composing modules: for each datum, modules are combined at each layer according to their normalized scores without requiring a task's ID (\S\ref{sec:lmc}). The local component is also used for module expansion: a new module is instantiated if all the current modules flag their input as being locally out-of-distribution (\S\ref{sec:expansion_strategy}). 
Further, new shallow modules (i.e. closer to the input) are first trained in a \emph{projection phase} to maximize the relatedness scores of subsequent, deeper, modules (\S\ref{sec:training}). This process projects the output of new modules into the representation space expected by the subsequent modules and ensures the compatibility between low- and high-level modules. 

In a set of studies, we explore the performance and versatility of our local structural approach, which we call Local Module Composer (LMC). First, we show that LMC reaches superior or comparable performance to existing modular and non-modular methods without requiring task IDs at test time using the Continual Transfer Learning (CTrL) benchmark, designed to evaluate transfer and forgetting in CL~\cite{veniat2020efficient} (\S\ref{sec:ctrl}). Then, we demonstrate how LMC, relying on its projection phase, can solve out-of-distribution (OOD) tasks not seen during the continual training (\S\ref{sec:c_ood}). We also show it is possible to combine modules from independently trained models into a new model to solve tasks seen by each of the independent models without any finetuning (\S\ref{sec:pnp}). Finally, an analysis of longer task sequences (30 and 100 tasks) reveals that LMC tends to spawn much fewer modules to reach good performance than the fully task-aware MNTDP~\cite{veniat2020efficient} counterpart. However, LMC reaches slightly lower accuracy on longer sequences than MNTDP, which highlights the difficulty of automatic task-ID agnostic module selection in the presence of a large number of candidate modules. In Appendix~\ref{app:meta-cl} we demonstrate the applicability of LMC in the meta-continual learning (meta-CL) setting, a task-agnostic setting by nature. %

We highlight that by relying on a local (per-module) structural component, LMC offers a modular CL approach that i) does not require task IDs during test in the standard task incremental settings; ii) balances parameter sharing to yield strong CL performances compared to baselines that require access to the task ID; iii) in our experiments instantiates a sub-linear number of modules; iv) permits recombination of modules at test time enabling OOD generalization as well as (v) the ability to combine independently trained models in a third model without fine-tuning. Notably, the OOD generalization is only possible if the agent is task-agnostic in the module selection process, since OOD tasks were not observed at training, the learner has to interpolate between the learned tasks, and a (categorial) task ID is of no use.
\begin{figure}[t]
    \centering     
    \includegraphics[width=0.83\linewidth]{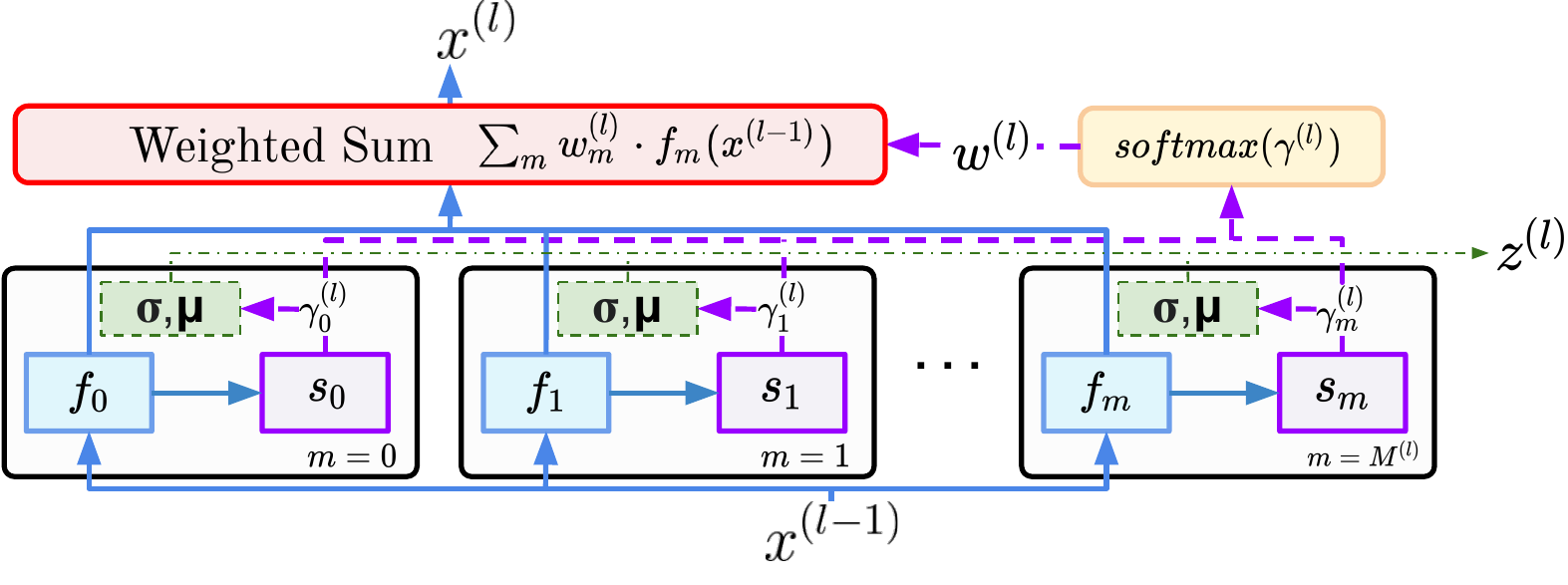}   
    \caption{\textbf{Modular Layer Scheme}. Each black rectangle is a module. Inside each module, the functional component $f_m$ receives the input $x^{(l-1)}$ and feeds its output to the structural component $s$. The output of the structural component is used to calculate the importance score $\gamma_m^{(l)}$ using Eq.~\ref{eq:gamma_l}, which are normalized to obtain the attention vector $w^{(l)}$. The layer output is the weighted sum of the functional outputs of each module. $\mu$ and $\sigma$ are the running mean and variance of the scores $\gamma_m^{(l)}$ used to detect outlier inputs and to trigger module addition.}
    \label{fig:fig_1} 
    \vspace{-1em}
\end{figure}        

\section{Background: Modular Continual Learning} 
\label{sec:background}
Let $\mathcal{F}(x;\theta): \mathcal{X} \rightarrow \mathcal{Y}$ be a learner parametrized with a set of parameters $\theta$. In task-incremental CL, the learner is exposed to a sequence of tasks. Each task is composed of a training set $D_t$ of $(x,y)$ pairs and a task identifier (ID) $t$ %
\cite{veniat2020efficient,kirkpatrick2017overcoming}. %
The goal is to learn an optimal $\theta^*$ that minimizes the loss $\mathcal{L}$ for all observed tasks:
\begin{align}
    \theta^* = \argmin_\theta \sum_{t=1}^T \mathbb{E}_{(x,y)\sim D_t} [\mathcal{L}(\mathcal{F}(x;\theta), y)].
\end{align}

The parameter sharing trade-off between tasks can be addressed through different architectural design choices for $\mathcal{F}$. For example, $\mathcal{F}$ can be a monolithic network that shares parameters $\theta$ across all tasks. Most existing task incremental CL methods use a task-specific output head, requiring the task ID to select the output head corresponding to the task at hand \cite{kirkpatrick2017overcoming,serra2018overcoming,Aljundi17}. 

At the other end of the spectrum are the expert based solutions that learn an independent model, a.k.a. expert, for each task \cite{aljundi2017expert,rusu2016progressive}. In this case, each expert trains task-specific parameters $\theta=\{\theta^{(t)}\}_{t=1}^T$. 

To balance parameter sharing and transfer, modular methods
organize their parameters in a series of modules $M = \{m_k^{(l)}\}$ with parameters $\theta = \{\theta_k^{(l)}\}$, where $\theta_k^{(l)}$ denotes the parameters of module $k$ at layer $l$ in $\mathcal{F}$. In general, a module can be any parametric function. In our experiments, unless otherwise stated, a module consists of a single convolutional layer followed by batch-norm, ReLU activation, and a max-pooling operation. 

Modules can be composed conditioned on a sample, a batch of samples, or a task. Let $\psi$ denote a specific composition of modules that gives rise to a distinct prediction function; we make this dependence explicit: $\mathcal{F}(x;\theta,\psi)$. Importantly, sharing modules across tasks should lead to desirable transfer properties.

\citet{veniat2020efficient} frames modular CL as finding an optimal layout $\psi^{(t)}$ for each task, where each layout selects a single module per layer per task (hard selection):
\begin{align}
    \theta^*&, \Psi^* = \argmin_{\theta, \Psi} \sum_{t=1}^T \mathbb{E}_{(x,y)\sim D_t} [ \mathcal{L}(\mathcal{F}(x ;\theta, \psi^{(t)}), y) ].
\end{align}
In this case the set of layouts $\Psi = \{\psi^{(t)}\}$ grows with the number of tasks, while modules can be reused across different task-specific layouts resulting in sub-linear growth pattern. They design a method called MNTDP to search the exponentially large space of modular layouts by only considering layouts resulting from adding a new module per layer to the best prior path (past task's path with the highest nearest neighbor accuracy on a new task) starting at the top layer. This solution relies on task IDs to retrieve $\psi^{(t)}$ at test time. %

Another way of composing modules uses dynamic routing \cite{mendez2020lifelong,rosenbaum2019routing,kirsch2018modular,meyerson2017beyond}.
The module layout is generated by a structural function $\psi = s(x)$, hence different inputs take different routes through $\mathcal{F}$. It is standard to approximate the structural function using a neural network $\psi = s(x;\phi)$ with structural parameters $\phi$. This framework has been applied to CL in~\cite{mendez2020lifelong} by learning a separate structural function per task $\psi^{(t)} = s(x;\phi^{(t)})$. The task IDs are used to retrieve the correct structural function:
\begin{align}
    \theta^*&, \Phi^* = \argmin_{\theta, \Phi} \sum_{t=1}^T \mathbb{E}_{(x,y)\sim D_t} [ \mathcal{L}(\mathcal{F}(x; \theta, s(x;\phi^{(t)})), y)],
\end{align}
where $\Phi = \{\phi^{(t)}\}$ is the set of structural parameters for all tasks.

The above methods require task IDs at both training and testing time. Next we introduce our modular CL approach that only relies on task IDs during training.

\section{Local Module Composer (LMC)}\label{sec:lmc}
\begin{figure}[t]
    \centering     
    \includegraphics[width=1\linewidth]{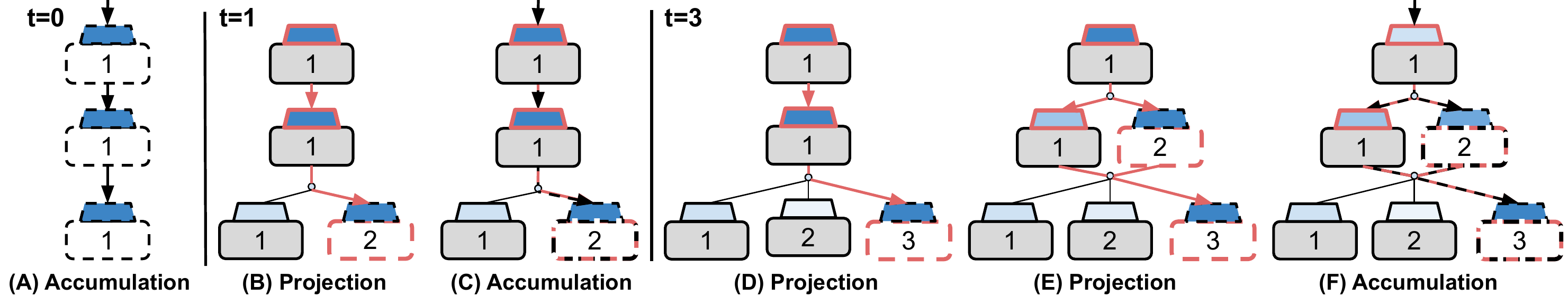}    
    \caption{\textbf{Two-phase training}. Each module contains a functional (rectangle) and structural (trapezoid) component. Their color intensity denotes the strength of their activation. Components with dashed contours are trained, solid contours represent fixed components, arrows show the gradient flow: black arrow --- \textit{functional} signal, pink --- \textit{structural} signal. \textbf{(A)} All modules are trained on task 0. \textbf{(B)} Task 1 arrives, a new module is added at layer 1, which is first trained to project its output into the representation of the modules above via the \textit{structural} signal (the \textit{functional} signal is optional). No module addition is allowed during the projection phase. \textbf{(C)} Module addition is allowed again, both signals are used for training. \textbf{(D)} As task 3 arrives, a new module is added at layer 1 again, projection phase is triggered. \textbf{(E)} A new module is added at the layer 2, both new modules are now trained in the second projection phase. \textbf{(F)} Both new modules are trained using both signals.}
    \label{fig:schematic}
\end{figure}
We propose LMC, a modular approach where each module consists of a functional component $f(x;\theta_m^{(l)})$ and a structural component $s(x;\phi_m^{(l)})$, see Figure~\ref{fig:fig_1}. The functional components are responsible for learning to solve the prediction task and are trained via the usual task loss $\mathcal{L}$ (e.g.\ cross-entropy loss for classification). The structural components receive the corresponding functional output as their input (see Figure~\ref{fig:fig_1}) and are responsible for dynamic routing through $\mathcal{F}$. Structural parameters $\phi$ are trained using a structural loss $\mathcal{L}^{(st.)}$ computed locally at each module. 

Intuitively, the structural component of a module should serve as a density estimator of the outputs of the functional component. The module's contribution to the layer's output is proportional to the likelihood of the input sample under the estimated density. In our instantiation, the structural component produces a relatedness score: a lower score for inputs that are more likely to belong to the distribution on which a given module was trained, and a higher score for inputs that are out-of-distribution for the given module. Hence, the likelihood of the input sample is approximated by the negative relatedness score.

Given an input data sample $x^{(0)}=x$, the output $x^{(l)}$ of a layer $l$ is defined as the weighted sum of the functional outputs of all $|M^{(l)}|$ local modules and used as input to the subsequent layer $l+1$:
\begin{equation}
x^{(l)} = \sum_{m=1}^{|M^{(l)}|} w^{(l)}_m \cdot f( x^{(l-1)};\theta^{(l)}_m)
\label{eq:x_l}.
\end{equation}

The functional output of the network is equal to the output of the final layer: $\mathcal{F}(x;\theta,\phi)=x^{(L)}$. In the last layer $\mathcal{F}$ implements a single output-head per task. At training time, the task ID is available and we update only the output-head corresponding to the currently learned task. At test time, the task ID is not available and we select the output head with the highest activation weight $w^{(L)}_m$, i.e.,\ the last layer performs hard module selection.

The module activation weights $w^{(l)}_m$ are computed by normalizing the vector of local relatedness scores $\gamma^{(l)}\in \mathbb{R}^{|M^{(l)}|}$. Each element of $\gamma^{(l)}$ is obtained from the negative structural loss which approximates the likelihood of each module:
\begin{align}
 \gamma_m^{(l)} &= - \mathcal{L}^{(st.)}\Big(s\big[f(x^{(l-1)};\theta^{(l)}_m);\phi^{(l)}_m\big]\Big) \label{eq:gamma_l}, \\
 w^{(l)}_m &= \text{softmax}(\gamma^{(l)})_m.\label{eq:w_l}
\end{align}
Modules with lower structural loss get higher activation weights. Note that in practice, it can be useful to bias the module selection towards the expected module selection in a batch, assuming that samples within a batch are likely to belong to the same task. We discuss this point further in \S~\ref{app:batched_module_selection}.

Instead of using the softmax function, it is possible perform hard selection taking the module with the highest score \citep{rosenbaum2019routing}, or alternatively selecting top-k modules \citep{shazeer2017outrageously}. In both cases, LMC's structural parameters stay differentiable due to the local nature of structural learning. Note that in the case of global structural objective, hard module selection would require applying tools for non-differentiable learning such as Expectation Maximization~\cite{kirsch2018modular} or reinforcement-learning based methods~\cite{rosenbaum2019routing,andreas2016learning}.

The overall LMC objective consists of optimizing both functional and structural losses:
\begin{align}
    \label{eq:overall_objective}
    \theta^*, \phi^* & = \argmin_{\theta, \phi} \sum_{t=1}^T \mathbb{E}_{(x,y)\sim D_t} \Big[ \mathcal{L}\big(\mathcal{F}(x; \theta, \phi), y\big) + \sum_{l=0}^L \sum_{m=0}^{|M^{(l)}|} \mathcal{L}^{(st.)}_m\big(s[f(x^{(l-1)};\theta^{(l)}_m);\phi^{(l)}_m]\big)\Big].
\end{align}

As in \cite{veniat2020efficient}, learning is performed w.r.t. only newly introduced modules to prevent forgetting.  

\paragraph{Structural component.} We test two instantiations of the structural component $s$ and loss $\mathcal{L}^{(st.)}_m$. In the first one, $s$ is an invertible neural network \citep{rezende2015variational}. Here we use the invertible architecture proposed by \citet{dinh2014nice}. As shown by \citet{hocquet2020ova}, for this invertible architecture the structural objective can be defined as $\mathcal{L}^{(st.)}_m(x)=||x||_2$. Intuitively, an invertible architecture prevents $\mathcal{L}^{(st.)}_m$ from collapsing to an all-zeros solution. 

In the second instantiation, $s$ and $f$ form an autoencoder and $\mathcal{L}^{(st.)}_m(x)=||x^{(l-1)} - x||^2$ is the reconstruction error with respect to the module's input $x^{(l-1)}$. \citet{aljundi2017expert} used a similar idea was for selecting the most relevant expert network conditioned on a task. Unless stated otherwise, modules in the feature extractor use the autoencoder as their structural component, while output heads use invertible $s$ --- these combinations worked well in practice. %

\subsection{Expansion strategy}           
\label{sec:expansion_strategy}                          
It is necessary to expand $\mathcal{F}$ as new tasks arrive to acquire new knowledge. A new module is added to a layer when all modules in this layer detect an outlier input. To this end, we track the running statistics of the relatedness score $\gamma$ for each module --- mean $\mu$ and variance $\sigma$ (see Figure \ref{fig:fig_1}), and calculate a z-score for each sample in the batch and each module at a layer:
\begin{align}
    z_{m} = \frac{w_{m} - \mu_m}{\sigma_m} \label{eq:z_score}.
\end{align}
An input is considered an outlier if its z-score is larger than a predefined threshold $z^{\prime}$ (see Appendix~\ref{app:ablation_threshold} for an ablation study of $z^{\prime}$ values). The expansion decision can be made on the per-sample (i.e.,\ if an outlier sample is detected) or a per-batch basis (i.e.,\ $z$ is averaged over the mini-batch). Unless stated otherwise, in our experiments, the decision was made on a per-batch basis. Additionally, at training the parameters of existing modules are fixed once the task changes. If during a forward pass through $\mathcal{F}$ module addition is triggered at multiple layers, we start adding modules at the layer closest to the input.

\subsection{Training} \label{sec:training}
Each module in LMC receives two types of learning signal: a structural signal resulting from minimizing $\mathcal{L}^{(st.)}_m$, and a functional signal resulting from minimizing the global functional loss $\mathcal{L}$. All structural components $s$ are trained \textit{only} with the structural signal that is calculated locally to each module. 

The training of functional components proceeds in two phases: projection and accumulation. Whenever the expansion strategy triggers the addition of a new module (i.e.,\ $z_m>z^{\prime}$ $\forall m\in \{0,\dots,|M^{(l)}|\}$), starting with layers closest to the input, LMC initiates the \textit{projection phase}.  During this phase, the new module is trained to minimize the structural loss from all the layers above and \textit{no new-module addition is allowed}. This procedure makes the representation of new modules compatible with subsequent modules and enables their composition. This procedure ``encourages'' already-learned modules to be reused, preventing over-spawning new modules. The functional signal is optional during projection (we kept it in all experiments unless otherwise stated). 

In the \emph{accumulation phase}, new module addition is allowed again and all non-frozen modules are trained with both signals. The functional components of new modules still receive a signal from  the structural components of modules above. The two-phase training is explained schematically in Figure~\ref{fig:schematic} and implementation details are provided in Appendix~\ref{app:implementation_details}.

\section{Experiments}
We now evaluate the performance, empirical capabilities, and properties of LMC in four different CL settings. First, in \S~\ref{sec:ctrl} we study a standard task-incremental CL setting (task-ID agnostic and aware) using the Continual Transfer Learning Benchmark (CTrL) \cite{veniat2020efficient}. Next, we explore the properties of LMC through other CL settings. In \S~\ref{sec:c_ood} we evaluate the continual OOD generalization ability of the proposed LMC. In \S~\ref{sec:pnp} we show the ability of LMC to combine modules form independently trained models. In Appendix~\ref{app:meta-cl}, we evaluate LMC in the Continual Meta-Learning setting.

\subsection{Continual transfer learning using the CTrL benchmark}
\label{sec:ctrl}

\newcommand{\ttiny}[1]{{\fontsize{5}{5}\selectfont #1}}
\newcommand{\pmt}{\tiny\textpm}
\newcolumntype{g}{>{\columncolor{Gray}}c}
\newcolumntype{d}{>{\columncolor{Gray}}l}
\setlength{\tabcolsep}{0.8pt}
\renewcommand{\arraystretch}{1}

\begin{table*}[t]
  \centering
  \centerline{    
  \begin{scriptsize}
  \begin{sc}             
  \renewcommand{\arraystretch}{1.15}
    \begin{tabular}{l|gcg|cgc|gcg|cgc|gcg}
    \toprule
    & \multicolumn{3}{c|}{$S^-$} & \multicolumn{3}{c|}{$S^+$} & \multicolumn{3}{c|}{$S^{in}$} & \multicolumn{3}{c|}{$S^{out}$} & \multicolumn{3}{c}{$S^{pl}$}  \\
    Model &  $\mathcal{A}$ & $\mathcal{F}$ & M & $\mathcal{A}$ & $\mathcal{F}$ & M & $\mathcal{A}$ & $\mathcal{F}$ & M & $\mathcal{A}$ & $\mathcal{F}$ & M & $\mathcal{A}$ & $\mathcal{F}$ & M \\ 
    \midrule 
        HAT\tiny\citep{serra2018overcoming}  & 63.7\pmt0.7 & -1.3\pmt0.6 & 24$^*$ & 61.4\pmt0.5 & -0.2\pmt0.2 & 24$^*$ & 50.1\pmt0.8 & 0.0\pmt0.1 & 24$^*$ & 61.9\pmt1.3 & -3.2\pmt1.3 & 24$^*$ & 61.2\pmt0.7 & -0.1\pmt0.2 & 20$^*$  \\
        EWC\tiny\citep{kirkpatrick2017overcoming}  & 62.7\pmt0.7 &  -3.6\pmt0.9 & 24$^*$ & 53.4\pmt1.8 & -2.3\pmt0.4 & 24$^*$ & 56.3\pmt2.5 & -9.1\pmt3.3 & 24$^*$ & 62.5\pmt0.9 &  -3.6\pmt0.9 & 24$^*$ & 54.2\pmt3.1 & -4.2\pmt2.7 & 20$^*$  \\
        O-EWC\tiny\citep{schwarz2018progress} & 62.0\pmt0.7 &  -3.2\pmt0.7 & 24$^*$ & 54.6\pmt0.7 & -1.3\pmt1.0 &  24$^*$ & 54.2\pmt3.1 & -10.8\pmt3.1 & 24$^*$ & 62.4\pmt0.6 &  -3.0\pmt0.9 & 24$^*$ & 52.3\pmt1.4 & -5.7\pmt1.3 & 20$^*$  \\   
        ER\tiny\citep{rolnick2019experience,chaudhry2019continual}  & 60.6\pmt0.7 &  -2.1\pmt0.9 & 4$^*$ & 63.0\pmt0.6 &  3.8\pmt0.8 & 4$^*$ & 63.8\pmt1.4 &  -1.9\pmt0.6 & 4$^*$ & 60.7\pmt1.0  &  -1.5\pmt0.5 & 4$^*$ & 60.5\pmt1.0 &  0.5\pmt0.9 & 4$^*$  \\
        \midrule 
        \midrule 
        Experts  & 62.7\pmt0.9 & 0.0 & 24 & \textbf{63.2\pmt0.8} & 0.0 & 24 & 63.1\pmt0.7& 0.0 & 24 & 63.1\pmt0.7 & 0.0 & 24 & \textbf{63.9\pmt0.5} & 0.0 &  20 \\
        MNTDP\tiny\citep{veniat2020efficient} & 66.3\pmt0.8  & 0.0 & 13.7 & 62.6\pmt0.7& 0.0 & 21.0 & 67.9\pmt0.9 & 0.0 & 16.0 & 65.8\pmt0.9 & 0.0 & 15.0 & 64.0\pmt0.2 & 0.0 & 17.2 \\
        SG-F\tiny\citep{mendez2020lifelong} & 63.6\pmt1.5 & 0.0& 14.7 & 61.5\pmt0.6 & 0.0& 20.8 & 65.5\pmt1.8 & 0.0& 17.5 & 64.1\pmt1.3 & 0.0& 16.2 &  62.0\pmt1.3 & 0.0& 16.0 \\
        LMC\tiny($\lnot$ A) & \textbf{66.6\pmt1.5}  & -0.0\pmt0.1 & 15.3 & 60.1\pmt2.7& -1.4\pmt2.4 & 21.3 & \textbf{69.5\pmt1.0} & 0.0\pmt0.1 & 20.0 & \textbf{66.7\pmt2.2} & -0.1\pmt0.1 & 15.5 & 61.6\pmt4.8 & -3.5\pmt3.1 & 18.2 \\
        \midrule
        MNTDP\tiny(A) & 41.9\pmt2.5 &-2.8\pmt0.6 &14.8 & 43.2\pmt1.3 & -10.8\pmt2.0& 20.7 & 32.7\pmt13.6 &-15.2\pmt13.2 &17.2 & 37.9\pmt2.7& -5.8\pmt3.5 &13.3 & 35.1\pmt3.6 & -16.4\pmt4.6 & 15.8 \\
        
        LMC\tiny(A) & %
        \textbf{67.2\pmt1.5} &  -0.5\pmt0.4 & 15.7 & 62.2\pmt4.5 &  2.3\pmt1.6 & 22.3 & \textbf{68.5\pmt1.7} & -0.1\pmt0.1 & 19.7 & \textbf{55.1\pmt3.4} & -7.1\pmt4.0 & 15.5 & \textbf{63.5\pmt1.9} & -1.0\pmt1.5 & 19.0 \\
          
        LMC\tiny(A,H) & 64.9\pmt1.5 & -0.2\pmt0.2& 16.2 & 55.8\pmt2.5 & -0.3\pmt1.2 & 15.3 & 67.6\pmt2.7 & -0.8\pmt1.0 & 21.5 & 54.2\pmt3.6 & -2.9\pmt2.0 & 15.9 & 53.8\pmt5.7 &3.1\pmt5.5& 10.8 \\
            
        SG-F\tiny(A) & 29.5\pmt3.5 & -35.3\pmt4.0 & 14.3 & 20.4\pmt4.4 &  -39.3\pmt6.7 & 16.0 & 24.4\pmt5.6 &  -38.7\pmt4.0 & 18.7 & 30.5\pmt4.5 & -34.0\pmt5.5 & 12.2 & 19.4\pmt1.0 & -41.8\pmt1.6 & 15.5 \\
        ER\tiny(A,S)\citep{rolnick2019experience,chaudhry2019continual} &  60.4\pmt1.0 &  -0.5\pmt0.7 & 4$^*$ & \textbf{65.3\pmt0.9} &  6.0\pmt1.0 & 4$^*$ & 58.8\pmt3.2 &  -4.2\pmt3.7 & 4$^*$ & 47.6\pmt1.5 &  -7.6\pmt1.6 & 4$^*$ & 58.6\pmt1.3 & -1.2\pmt1.5 & 4$^*$ \\
        \midrule
        \midrule 
         
        Finetune & 47.5\pmt1.5 &  -14.9\pmt1.4 & 4$^*$ & 31.4\pmt3.7 &  -29.3\pmt3.8 & 4$^*$ & 39.7\pmt5.0 &  -23.9\pmt5.7 & 4$^*$ & 45.4\pmt4.0  &  -15.5\pmt3.7 & 4$^*$ & 29.1\pmt3.1 & - 29.2\pmt3.2 & 4$^*$  \\
        
        Finetune L & 52.1\pmt1.4 &  -15.7\pmt1.7 & 24$^*$ & 38.2\pmt3.2 &  -25.8\pmt3.3 & 24$^*$ & 49.3\pmt2.0 &  -18.4\pmt2.0 & 24$^*$ & 49.3\pmt2.1  &  -18.4\pmt2.0 & 24$^*$ & 37.1\pmt2.1 &  -26.0\pmt2.2 & 20$^*$ \\ 
        \bottomrule
    \end{tabular}
  \end{sc}
  \end{scriptsize}
  }
  \caption{\textbf{CTrL results:} we report  accuracy ($\uparrow\mathcal{A}$), forgetting ($\uparrow\mathcal{F}$) with standard deviations calculated over 6 different runs. We report the mean number of modules (M) over these runs, where $^*$ marks methods with fixed capacity. The first block comprises a set of standard CL baselines including regularization and replay based methods. The second block are the modular methods, third -- modular and replay based methods that are task ID agnostic (A), and the last block are the two finetuning baselines. (H) indicates hard module selection. (S) indicates single-head as detailed in the main text.}
 \label{tab:tab1}
\end{table*}

The CTrL benchmark was proposed to systematically evaluate properties of CL methods with a focus on modular architectures \cite{veniat2020efficient}. It consists of 5 streams of visual image classification tasks. The first stream $S^-=(t_1^+,t_2,t_3,t_4,t_5,t_1)$ consists of a sequence of 6 tasks, where the first and last task are the same except the first has an order of magnitude more training samples (``$^+$'') than other tasks. This stream is designed to evaluate the \emph{direct transfer} ability of models, i.e.\ a modular learner should be able to \textit{reuse} the first task's modules for the last task. The $S^+=(t_1,t_2,t_3,t_4,t_5,t_1^+)$ stream is similar to $S^-$, but now the last task comes with more data than the other tasks (including the first one). Here, the modular learner should be able to \emph{update its knowledge}, i.e.\ performance on the first task should improve after learning the last task. In the $S^{in}=(t_1,t_2,t_3,t_4,t_5,t_1^{\prime})$ stream the first $t_1$ and the last $t_1^{\prime}$ tasks are similar, with a slight input distribution change (e.g.\ different background color). In the $S^{out}=(t_1^+,t_2,t_3,t_4,t_5,t_1^{\prime\prime})$ stream the first task $t_1$ and the last task $t_1^{\prime\prime}$ differ in the amount of training data and the output distribution, i.e.\ the labels of the last task are randomly permuted. %
The \emph{plasticity} stream $S^{pl}=(t_1,t_2,t_3,t_4,t_5)$ evaluates the ability to learn a stream of unrelated and potentially interfering tasks, i.e.,\ transfer from unrelated tasks can harm performance. Descriptive statistics for all datasets are in Appendix~\ref{app:ctrl_streams}.

We compare to several baselines.    
\textbf{Finetune}: trains a single model (wider model marked with \textbf{L}) for all tasks. \textbf{Experts}: trains a model per task. We also compare with the several recently proposed modular CL baselines, which achieve competitive results in CTrL and require task IDs at test time. \textbf{MNTDP}~\cite{veniat2020efficient}: a recent search-based module selection approach described in more detail in \S~\ref{sec:background}. MNTDP requires the task ID to retrieve the previously found best structure for each test task. \textbf{MNTDP\tiny(A)}: a task ID agnostic version of MNTDP we created, which selects the path with the lowest entropy in the output distribution.~
\textbf{SG-F}~\citep{mendez2020lifelong}: Soft-gating with fixed modules, a modular method mentioned in \S~\ref{sec:background}. It relies on a task-specific structural network that generates soft-gating vectors for each layer of the modular learner and fixes learned modules when new tasks arrive. We slightly adapted the original expansion strategy of SG-F in order to conform to our experimental setup; details are in Appendix~\ref{app:baseline_details}. \textbf{SG-F\tiny(A)}: a version of SG-F with a single structural network shared across tasks. \textbf{HAT}\citep{serra2018overcoming}: learns attention masks for activations that gate the gradients to prevent forgetting. The task ID is used to select a task-specific attention mask for inference. 

We also compare to several standard CL methods.  \textbf{EWC}~\cite{kirkpatrick2017overcoming}: trains a single model for all tasks while applying parameter-regularization to minimize forgetting. \textbf{O-EWC}:\cite{schwarz2018progress} online version of EWC that does not require storing a separate approximation of the Fisher information matrix per task. \textbf{ER}~\cite{chaudhry2019continual}: trains a single model while replaying samples from previously seen tasks. The size of the replay buffer corresponds to the memory size of the LMC assuming the worse case linear growth pattern (i.e., LMC with 24 modules on a 6-task sequence). \textbf{ER\tiny(A,S)}: task ID agnostic version of ER that uses a single output head to classify all classes from all tasks: i.e.\ after learning stream $S^-$ the output head has 50 output neurons and the output classes of the last task are considered the same as the ones of the first task.  

We use several versions of LMC. \textbf{LMC{\tiny($\lnot$A)}} a version of LMC that uses the task ID for output head selection (not module selection as MNTDP). \textbf{LMC\tiny(A)}: the default version of LMC. It equips output heads with structural components and is therefore task ID agnostic at test time. \textbf{LMC\tiny(A,H)}: a version of task ID agnostic LMC that performs hard module selection, i.e.,\ taking the module with the highest relevance score per layer. %
All methods use the same architecture (described in Appendix~\ref{app:architecture}) together with the Adam~\citep{adam2015} optimizer. HAT is the only method that uses SGD.
 
Similar to \citet{veniat2020efficient}, we use the following evaluation metrics:  ($\mathcal{A}$) \textbf{average accuracy} on all seen tasks at the end of CL training; \textbf{Forgetting} ($\mathcal{F}$) --- difference between accuracy at the end of the training and accuracy after learning the task averaged across tasks~\cite{lopez2017gradient}; \textbf{Number of modules} (M) at the end of the continual training procedure. Formal definitions of all metrics are in Appendix~\ref{app:ctrl_metrics}.

Table~\ref{tab:tab1} reports performance using the CTrL benchmark. Overall, modular methods tend to outperform ER and the regularization based methods (HAT and EWC). Among the modular methods, soft-gating SG-F{\tiny(A,F)} with a single controller shared among all the tasks performed the worst. This baseline showcases the problem of forgetting in the global structural component (a.k.a. controller) of dynamic routing methods such as the one proposed by \citet{mendez2020lifelong}. 
A version of LMC performs the best on the $S^-$, $S^{in}$ and $S^{out}$ streams. 

Notably, LMC{\tiny(A)}, which does not rely on task IDs at test time, outperformed all other task ID agnostic methods such as MNTDP{\tiny(A)} and ER{\tiny(A,S)} on all streams but $S^+$, and always performed on par with task ID aware methods. On the $S^{out}$ stream low performance is expected for task-ID agnostic methods due to output distribution shift: i.e.,\ at test time we notice that LMC correctly assigns samples from the last task $t_1^{\prime\prime}$ to the first task's $t_1^+$ output head. However, the resulting classification accuracy is low because the labels of the last task are randomly permuted in this stream. 

The task ID agnostic LMC{\tiny(A)} outperforms task ID aware LMC on the $S^{-}$ and $S^{+}$ streams. Here, LMC{\tiny(A)} selects modules (and the output head) which were predominantly trained on the task that provided more training data (e.g.\ $t_1^+$ in $S^-$ stream), hence transferring knowledge between the first and the last tasks. In contrast, LMC{\tiny($\lnot$A)} when tested on the last task $t_1$ is forced to select the output head belonging to this task, which was trained on less data than the output head of $t_1^+$ task, leading to lower accuracy. In addition, we observed that versions of LMC often exhibit high variance (e.g. see $S^+$, $S^{out}$ and $S^{pl}$ streams). This may be caused by the larger amount of trainable parameters compared to other models and relatively small amount of training data. Finally, low performance of LMC{\tiny(A,H)} emphasizes the importance of soft modular attention for LMC. Additional results, including a transfer metric~\cite{veniat2020efficient}, are in Appendix~\ref{app:transfer_ctrl}.
    
\begin{figure}[!t]%
    \centering
    \begin{subfigure}[t]{0.24\textwidth}
    \centering
        \includegraphics[height=1.07\textwidth]{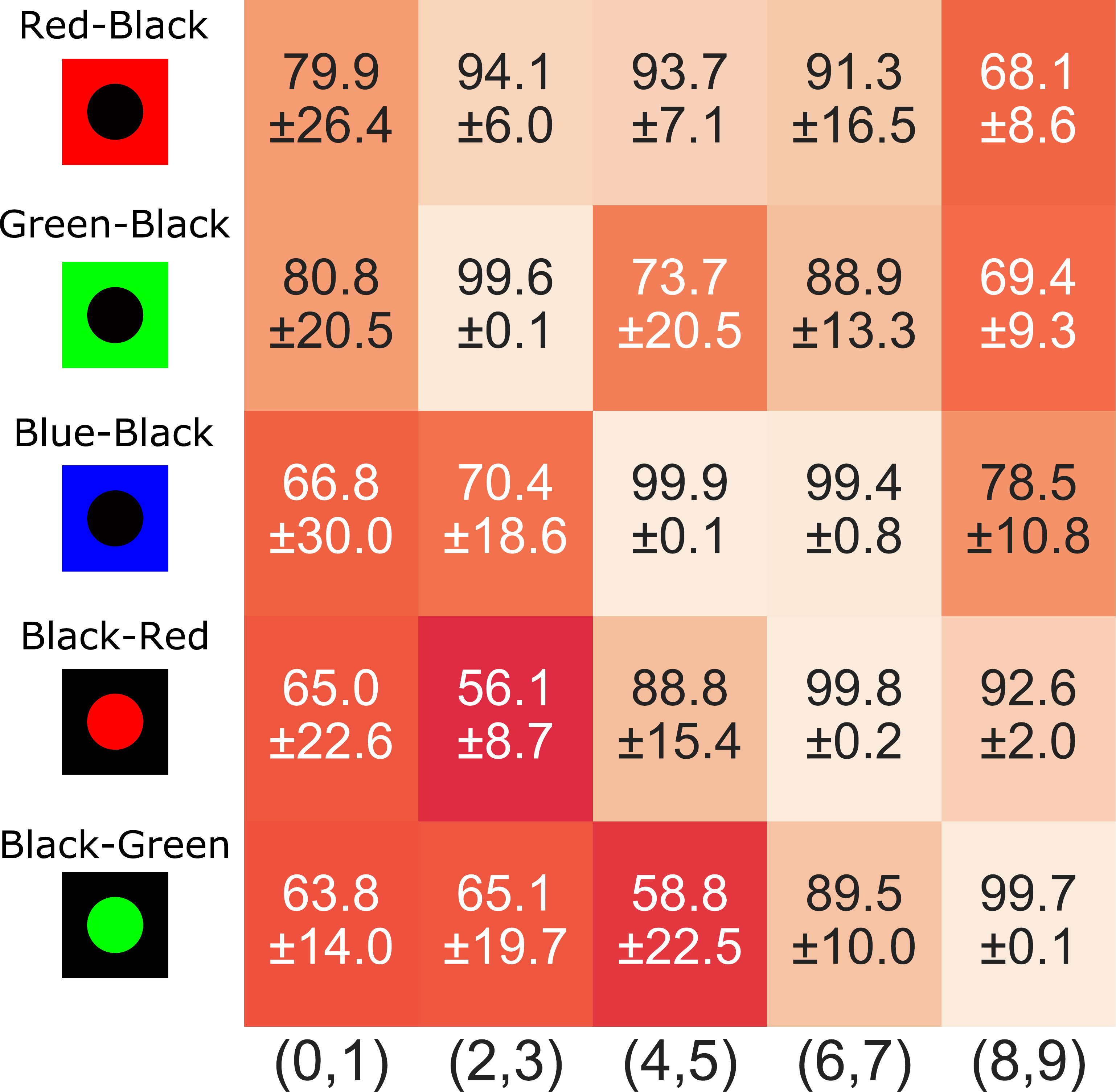}
        \caption{EWC}
        \label{fig:ood-ewc}
    \end{subfigure}
    \begin{subfigure}[t]{0.24\textwidth}
    \centering
    \hfill
        \includegraphics[height=1.07\textwidth]{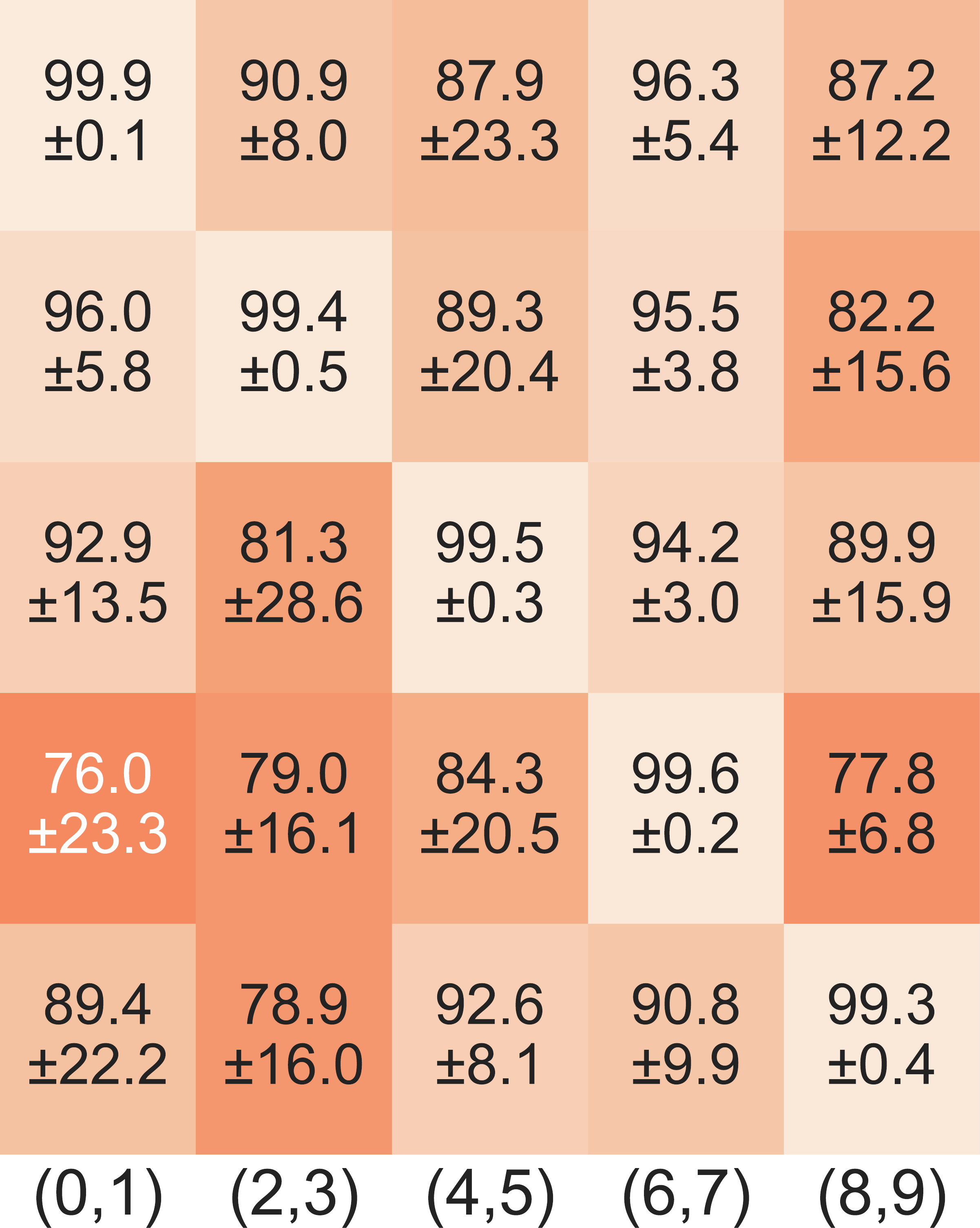}
        \caption{MNTDP}
        \label{fig:ood-mntdp}
    \end{subfigure}
    \begin{subfigure}[t]{0.24\textwidth}
    \centering
        \includegraphics[height=1.07\textwidth]{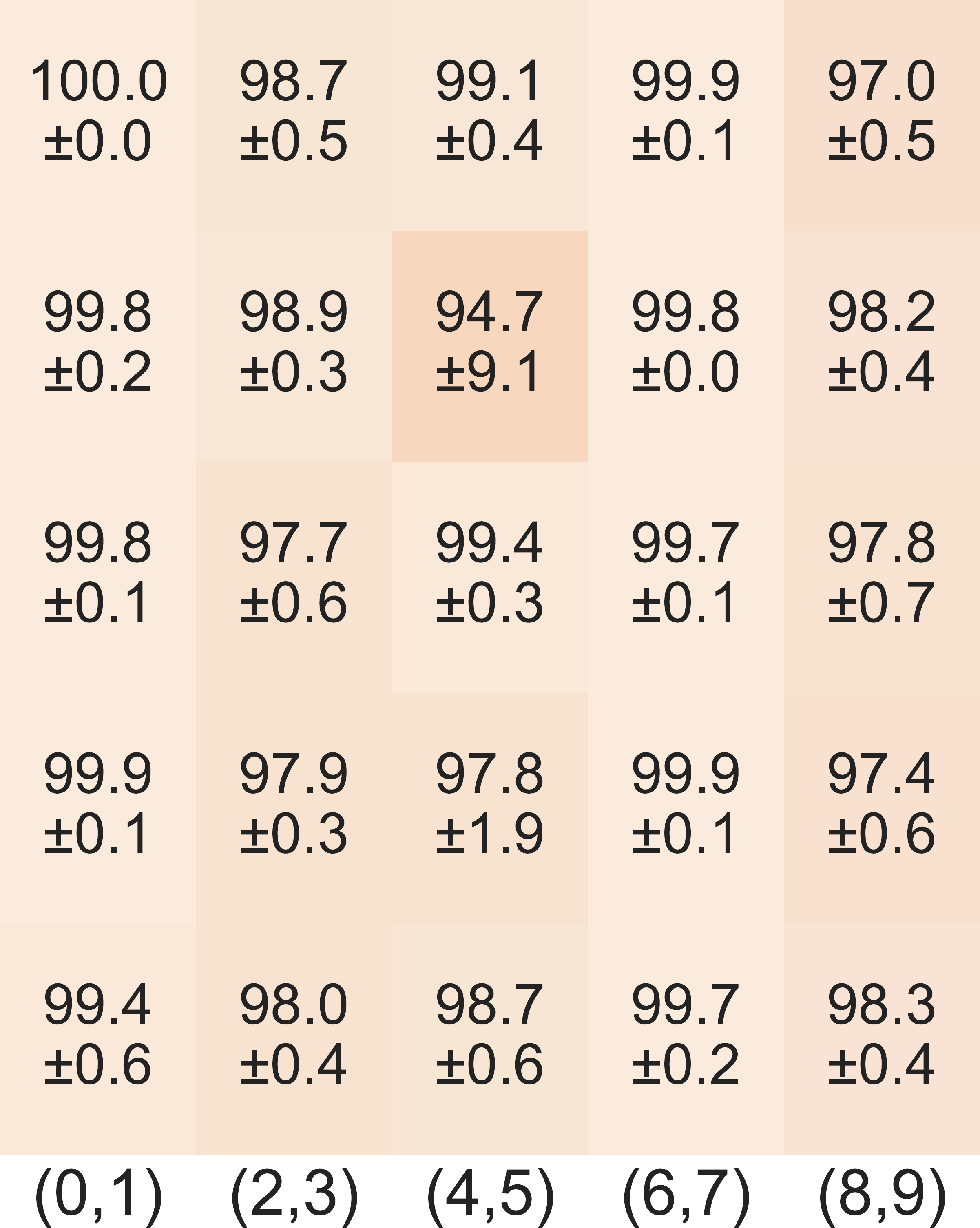}
        \caption{LMC{\tiny($\lnot$A)}}
        \label{fig:ood-lmc}
    \end{subfigure}
    \begin{subfigure}[t]{0.24\textwidth}   
    \centering
        \includegraphics[height=1.08\textwidth]{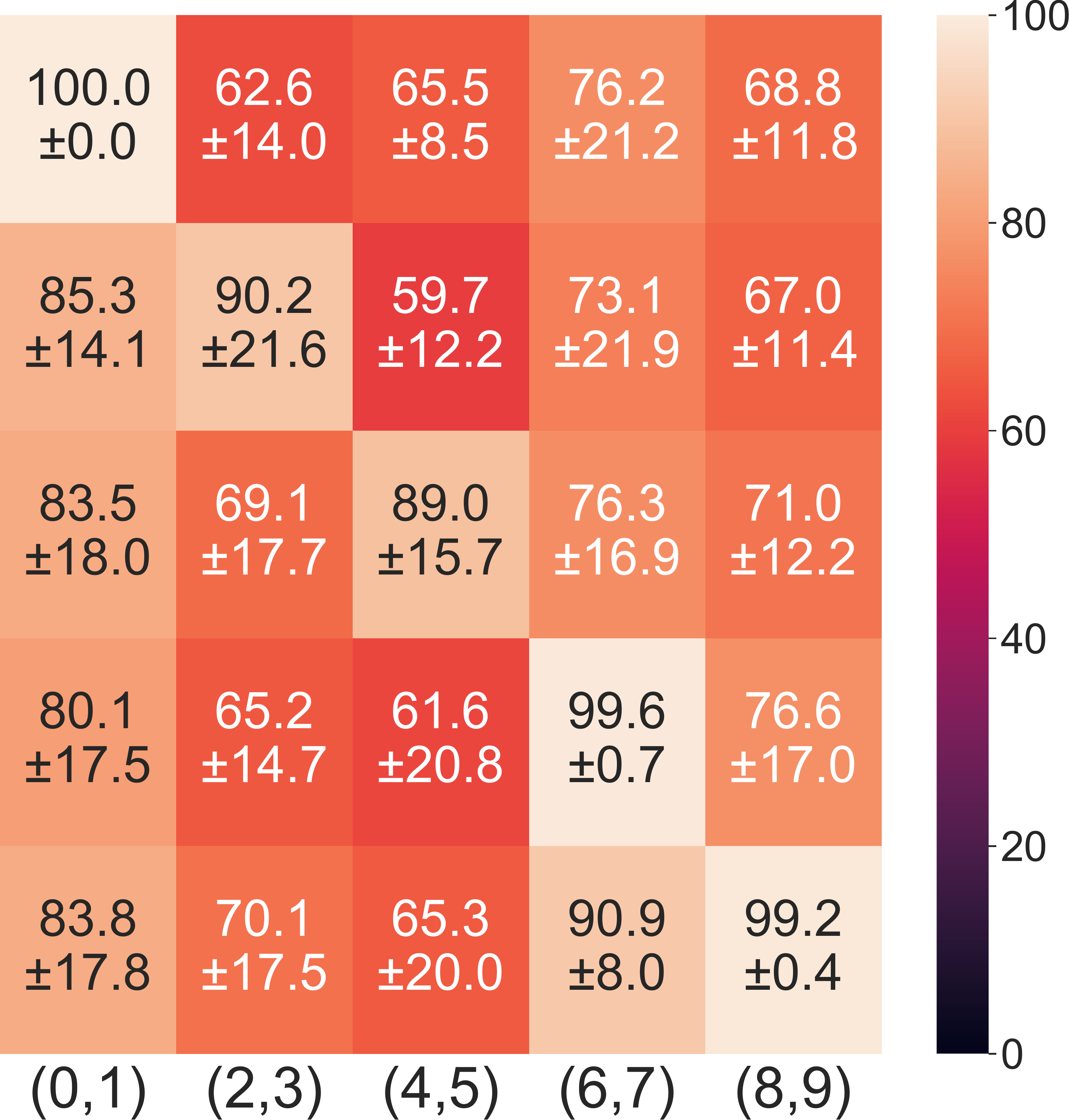}
        \caption{LMC{\tiny($\lnot$A)} (no proj.)}
        \label{fig:ood-lmc_no_projection}
    \end{subfigure}
    \caption{Continual OOD-generalization: matrices show test accuracy on seen and unseen tasks (on- and off-diagonal tasks respectively). In each sub-figure x-axis shows the MNIST class-combination used to build the task, y-axis gives the foreground-background colors. Only diagonal tasks are learned continually. While non-modular EWC (a) and modular MNTDP (b) can prevent forgetting, LMC{\tiny($\lnot$A)} (c) is able to generalize to OOD tasks as well. LMC{\tiny($\lnot$A)} without projection phase performs poorly as shown in (d).}%
    \label{fig:ood_confusion}
\end{figure}
\subsection{Compositional OOD generalization}
\label{sec:c_ood}
This second study tests the ability of LMC to recombine modules for OOD generalization. We use a colored-MNIST dataset --- a variation of the standard MNIST dataset of hand-written digits from 0 to 9~\cite{kim2019learning} in which digits are colorized. We design a simple sequence of tasks as follows. First, we define two high-level features: the foreground-background color combination (using the colors red, black, green, blue) and the class (0--9). Then, we create five non-overlapping tasks of two (digit) classes each: \{\texttt{0--1, ..., 8--9}\}. At training time the model is continually trained using a sequence of these tasks, however, each task is only seen in one of five different foreground-background combinations \{\texttt{red-black, green-black, blue-black, black-red, black-green}\}. At test time we measure the generalization ability to seen and unseen combinations of classes and colors.

In Figure~\ref{fig:ood_confusion} we present the accuracy matrices for different learners when tested on all 25 combinations of colors and classes after it has been trained only on the 5 tasks on the diagonal. 

We compare the performance of LMC{\tiny($\lnot$A)} with EWC~\cite{kirkpatrick2017overcoming}, MNTDP~\citep{veniat2020efficient}, and an ablated version of LMC without the projection phase. We observe that the OOD accuracy attained by LMC is significantly higher than EWC and MNTDP. Since the model trained with EWC is monolithic, the digit-background color combinations are entangled with the digits' shape for each task, hindering OOD generalization. %
In contrast, modular approaches such as MNTDP and LMC learn a different module combination for each task. In contrast to MNTDP, LMC's module selection does not rely on task identifiers and each module is selected in a local manner based on its compatibility with the current input. This allows LMC to interpolate between previously seen tasks being able to dynamically compose existing modules to adapt to tasks that have not been seen at training. Because MNTDP's module selection relies on a database of task-specific structures found to be optimal for the corresponding task at training, this method must reuse the predefined module compositions based on task IDs. This forces MNTDP to use modules that were trained using a different color combination, and results in e.g.\ a $24\%$ accuracy drop with respect to LMC on \texttt{[0,1]}  when the foreground and background colors are inverted w.r.t. the seen combination. 

In Figure~\ref{fig:ood-lmc_no_projection} we report results for LMC without applying the projection phase. The projection phase adapts the representation of newly introduced modules to match the distribution expected by the subsequent modules. As expected, we found that skipping it severely degrades performance. This result validates the usefulness of the projection phase to achieve an efficient local module selection. In Appendix~\ref{app:ood} we plot the average module selection for all 25 test tasks, showing how modules are reused for the OOD tasks.

\subsection{Combining modular learners}
\label{sec:pnp}

\begin{wrapfigure}{r}{.5\textwidth}
    \begin{minipage}{\linewidth}
    \centering
    \includegraphics[width=\linewidth]{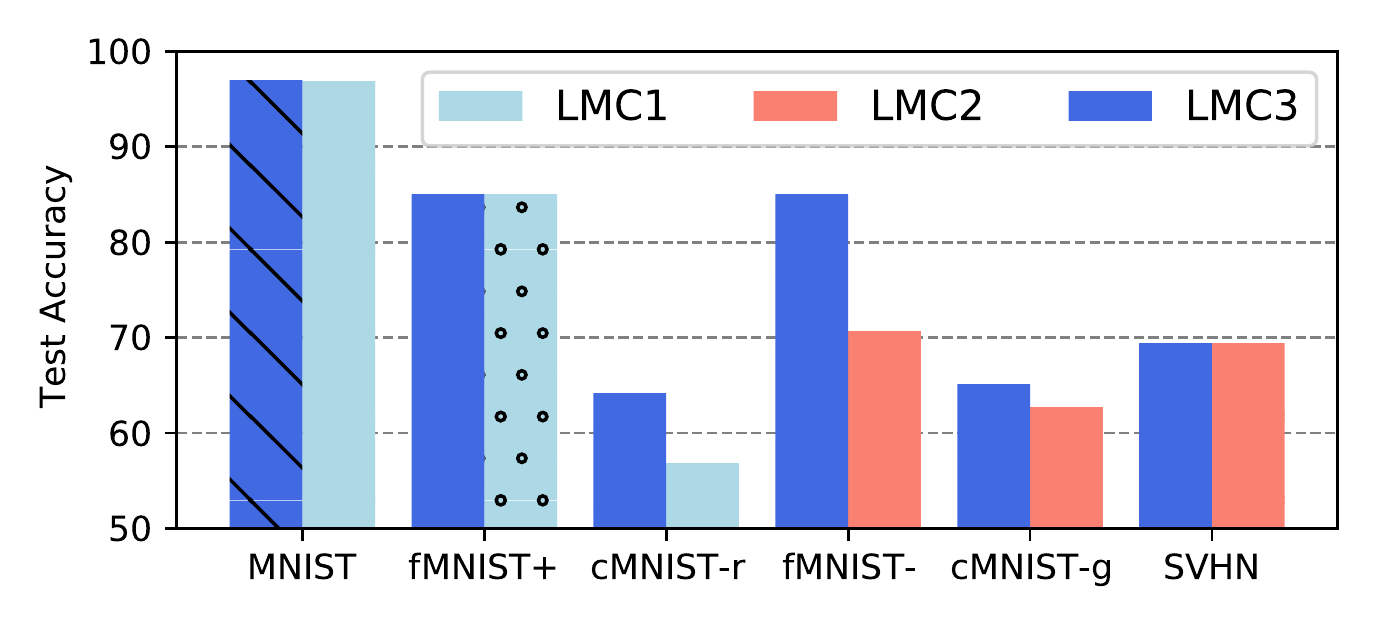}
\end{minipage}
\caption{Performance of combining independently trained LMC1 and LMC2 into LMC3.}
\label{fig:pnp}
\end{wrapfigure}

In earlier sections we show cross-task reusability of modules, here we test the cross-model reusability. We motivate the practical importance of this kind of reusability with a federated learning example: a privacy preserving training might be required for LMC1 and LMC2, trained on the premises of customers 1 and 2, after which their modules can be combined in a single central entity --- LMC3, located on premises of the cloud service provider. LMC3 is required to perform tasks seen by both independent LMCs but can not be finetuned as it has no access to the original training data distributions. %

In Figure~\ref{fig:pnp} we test the ability of LMC to preserve and transfer knowledge in such setting. To this end, we design the following tasks: fMNIST+ and fMNIST-. Both are sampled from the fashion-MNIST dataset \citep{xiao2017fashion} but the latter comes with an order of magnitude less training data.  cMNIST-\textbf{r} is a variant of the colored-MNIST dataset where the background of 95\% of the training samples is colored in \textbf{r}ed and 5\% in \textbf{g}reen. For the cMNIST-\textbf{g} dataset these proportions are inverted and 95\% of the training samples is colored in \textbf{g}reen. The test set contains 50\% of samples with green and 50\% with red background.  We trained LMC1 continually on MNIST, fMNIST+, and cMNIST-r tasks. We trained LMC2 on fMNIST-, cMNIST-g, and SVHN. We then combined the modules of both LMCs layer-wise to obtain LMC3.

We observed positive transfer for both cMNIST and fMNIST- tasks with LMC3. We found that LMC3 selects different modules originating from different LMCs conditioned on test samples with different background colors --- LMC1's modules were specialized on red background while LMC2's on green (selected paths presented in Appendix~\ref{app:combining_independent_models}). Notably, cross-model reusability without fine-tuning is novel to LMC and can be attributed to the local nature of the structural component. Using a global structural component as in \citep{mendez2020lifelong} would require tuning a separate structural component specifically for LMC3. In case of task-specific routing of proposed for MNTDP~\citep{veniat2020efficient}, additional search would be needed to discover task-specific paths through the consolidated LMC3 model. In both cases the access to the orinal training data distributions would be required and privacy would not be preserved.

\subsection{Longer task sequences}
\label{sec:longer_sequence}
\begin{figure}%
    \centering
    \subfloat[\centering 30 tasks sequence $S^{long30}$]{{\includegraphics[width=0.44\textwidth]{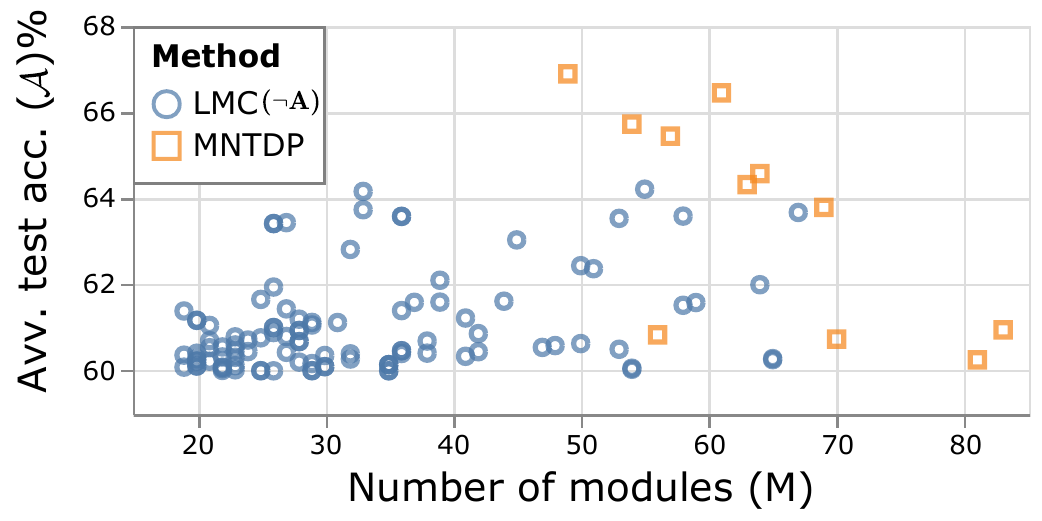} }}%
    \qquad
    \subfloat[\centering 100 tasks sequence $S^{long}$]{{\includegraphics[width=0.44\textwidth]{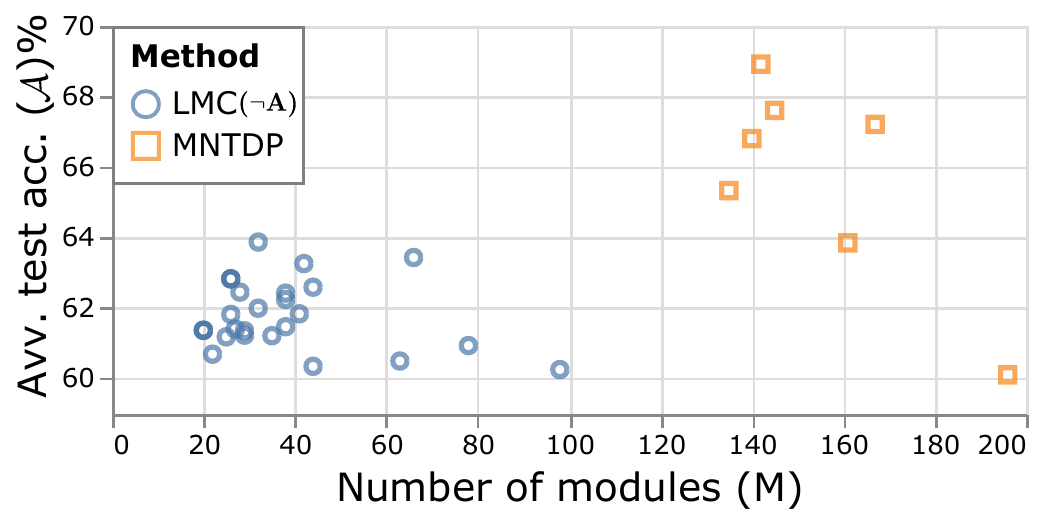} }}%
    \caption{Results on $S^{long}$ and $S^{long30}$ sequences for different hyperparameter values (we select only runs with reasonably good performance, i.e. $\mathcal{A}$>60\%), same plots plots for all conducted runs can be found in Appendix~\ref{app:long_modules_vs_acc}).}%
    \label{fig:acc_nmodules_slong}%
\end{figure}

Here we study the performance LMC on longer task sequences consisting of 30 -- $S^{long30}$, and 100 -- $S^{long}$ tasks. The $S^{long}$ sequence corresponds to the one proposed by~\citet{veniat2020efficient} as part of the CTrL benchmark. $S^{long30}$ is a 30-tasks subset of $S^{long}$ (see Appendix~\ref{app:ctrl_streams} for details). 

We first report the average test accuracy ($\mathcal{A}$) and the total number of modules (M) of the models selected through cross-validation. $S^{long30}$: MNTDP $\mathcal{A}$=64.58, M=64; {LMC\tiny($\lnot$A)}: $\mathcal{A}$=62.44, M=50. $S^{long}$: MNTDP $\mathcal{A}$=68.92, M=142; LMC{\tiny($\lnot$A)}: $\mathcal{A}$=63.88, M=32. While the gap between the accuracy of LMC and MNTDP on $S^{long30}$ is only $1.86$\%-points, in the case of $S^{long}$ this gap grows to $6.58$\%-points. It is important to highlight that in contrast to LMC, MNTDP's module selection is performed by a task ID aware oracle. We further analyze the trade-off between the number of modules and accuracy in Figure~\ref{fig:acc_nmodules_slong}, where we plot the number of modules (M) against average test accuracy ($\mathcal{A}$) for models that resulted from training with different hyperparameters. For both streams, we observe that LMC tends to spawn much fewer modules than MNTDP. However, MNTDP shines in the presence of large number of modules and achieves higher overall test accuracy on these streams. Interestingly, as can be clearly observed on the $S^{long}$ stream, LMC reaches higher accuracy with smaller number of modules: e.g.\ $\sim$64\% with 32 modules, while adding modules leads to lower accuracy: e.g.\ $\sim$61\% with 98 modules. This result suggests that local task ID agnostic module selection becomes more challenging for LMC in presence of a large number of modules.

\section{Related work} 

\textbf{Modularity in neural networks} is studied in the context of scalability~\cite{ballard1987modular}, and more recently as a way to achieve compositionality and systematic generalization~\cite{andreas2016neural,kirsch2018modular, chang2018automatically, bahdanau2018systematic, goyal2019recurrent, csordas2020neural} as well as for multi-task learning~\cite{meyerson2017beyond,rosenbaum2019routing}. 
From the causal point of view, a data generation process could be thought as a composition of independent causal modules~\cite{pearl2009causality}. Researchers model these kinds of systems using a set of independent modules, where each module is invariant to changes in the other modules induced by e.g.\ distribution shifts~\cite{scholkopf2012causal, peters2017elements}. This idea is crystallized by~\citet{parascandolo2018learning}, who propose a way to learn a set of causal independent mechanisms as mixture-of-experts. Building up on this ideas, others show evidence of compositional OOD generalization~\cite{madan2021fast}. Recently, \cite{mundt2020wholistic} argue for a more wholistic view on CL including OOD generalization as an important desiderata. %

\textbf{Continual learning} methods typically address the problem of forgetting through parameter regularization ~\cite{kirkpatrick2017overcoming,Nguyen17,zeno2018task}, replay~\cite{soltoggio2015short,rebuffi2017icarl, Aljundi2019Online,ostapenko2019learning,lesort2018generative, caccia2019online, wu2018memory,  isele2018selective, rolnick2019experience, caccia2021reducing} or dynamic architectures (and MoEs)~\cite{rusu2016progressive,serra2018overcoming,Dyn_expand_net_Lee,schwarz2018progress,lee2020neural,jerfel2019reconciling}. 
Our work falls under the umbrella of the latter and shares its advantage of having the capacity to adapt to a large number of related tasks. Our focus is on improving \textbf{modular CL} approaches, which despite their advantages, have only recently been studied in the CL literature~\cite{mendez2020lifelong,veniat2020efficient}. The main difference with our work is that we use a local composition mechanism instead of a global one. We detail this difference in \S\ref{sec:background} and also compare to these methods in \S\ref{sec:ctrl}. 

\textbf{Continual-meta learning} focuses on fast learning and remembering~\cite{finn2019online,he2019task,harrison2019continuous,jerfel2019reconciling}, often emphasising the online performance on OOD tasks~\cite{caccia2020online}. As argued by \citet{jerfel2019reconciling} modularity can be useful in this setting to minimize interference between tasks. They proposed a way to train a MoE model, with each expert focusing on a cluster of tasks leveraging Bayesian nonparametrics. LMC aims at decomposing knowledge into layer-wise composable modules further reducing modular granularity.
Continual-meta learning is often confused with its counterpart \emph{meta-continual learning} \citep{javed2019oml,caccia2021special,von2021learning}, in which algorithm are learning to continually learn.

The rapid growth of continual learning has lead researchers to work on empirical studies \citep{de2019continual,Lomonaco2020CVPR2C, lesort2021continual}, surveys \citep{hadsell2020embracing,khetarpal2020continual,lesort2021understanding, mundt2020wholistic,mundt2021clevacompass} as well as CL-specific software \citep{normandin2021sequoia, Douillard2021ContinuumSM, lomonaco2021avalanche}.

\section{Conclusion}                      
We develop LMC, a method to learn and compose a series of modules on a continual stream of tasks fulfilling some of the basic desiderata of modular CL such as module specialization, avoidance of collapse, and sublinear growth. In LMC, structural information is learned and stored locally for each module. It is the locality of the structural component that enables generalization to related but unseen tasks, and that permits combining different LMCs without fine-tuning.

Future work could focus on achieving more efficient sub-linear model growth through OOD generalization and reusability of modules. Additionally, while the benefits of modularity for CL are well understood, the implications of the  CL regime on modularity and compositionality have not been studied extensively. It is possible that providing knowledge to the learner in incremental chunks results in the implicit supervision needed to better disentangle it into specialized and composable modules. Another promising direction is removing the need for task boundaries during training and developing more robust architectures for the local structural component (related discussions are in Appendix \ref{app:structural_component}, and limitations in Appendix \ref{sec:limitations}).

{\small
\bibliographystyle{plainnat}
\bibliography{egbib}
}
\begin{ack}

Laurent Charlin holds a CIFAR AI Chair Program and acknowledges support from Samsung Electronics Co., Ldt., Google, and NSERC. Massimo Caccia was supported through MITACS during his part time employment with Element AI the ServiceNow company. Massimo Caccia was also supported by Amazon, during his part time employment there. We would like to thank Mila and Compute Canada for providing computational resources. We also would like to thank Irina Rish for useful discussions.

\end{ack}
\newpage

\newpage
\appendix

\section{Implementation and Algorithm details}
\label{app:implementation_details}               

\subsection{Batched modularity}
\label{app:batched_module_selection}
Activation weights $w_m^{(l)}$ can be calculated separately for each data point. However, in batched regimes sometimes it can be assumed that module selection for samples in the same batch is likely to be similar.\footnote{E.g in case of locally stationary data distribution, samples seen together are likely to belong to the same task} This can be incorporated into LMC by redefining $w^{(l)}_m$ as:
\begin{align}
&w^{(l)}_m =\frac{1}{Z(w)} \text{softmax}(\gamma^{(l)},\tau )_m \cdot \mu_{w,m}^{(l)}, \label{eq:w_l_batch}\\
&\mu_{w,m}^{(l)}=\mathbb{E}_{x_i\in b}\big(\text{softmax}(\gamma^{(l)}, \tau^\prime)_m\big),
\end{align}
where $Z(w)$ is the normalization term, $b$ denotes a batch of samples, $\tau$ and $\tau^\prime$ are the temperature hyperparameters. Lower $\tau^\prime$ would results in a stronger bias towards the expected selection $\mu_w^{(l)}$.

\subsection{Training with projection phase}
\label{app:projection_phase_details}                             
    The intuition behind the modular training with projection phase is the following: every time a new module addition is triggered (using the mechanism proposed in \S~\ref{sec:expansion_strategy}), we start by only adding modules on the deepest layer --- i.e.\ the one closest to the input. Then, in the \textit{projection phase}, we train this module for some time using the signal coming from the structural components of the modules above (possibly combined with the training signal of the downstream task). Projection phase makes sure that the learner first tries to efficiently \emph{reuse} the existing modules (the once above the newly added one) by trying to project it's output into the representation space expected by those modules. After some time the learner is allowed to add new modules again. If the previously added module was not enough to incorporate the distribution shift that caused the previous module addition, new module addition will be triggered in the layers above. We detail this procedure in the Algorithm~\ref{algo:lmc} and~\ref{algo:forward}. Additionally, for modules that recognize current input as outlier in the forward pass, we set their contribution for the current batch to zero (ll.14 in Algorithm~\ref{algo:forward}). This ensures that the newly added modules get enough training signal to learn.

\centerline{
\scalebox{0.85}{
    \begin{minipage}{0.55\linewidth}
        \setlength{\textfloatsep}{2mm}
        \begin{algorithm2e}[H]
            \small
            \SetAlgoLined
            \DontPrintSemicolon                 
            \textbf{Require:} $k$ projection phase length, stream $S$, $z$-score threshold $z^{\prime}$\\
            \textbf{Initialize} Learner $\mathcal{F}_{\theta,\phi}$ \\
            \For{ $t = 0 ... |S|$}{     
                $D_t \leftarrow$ get dataset for task $t$  \\
                \For{e = 0 ... total epochs }{  
                    \ForEach{ mini-batch $b \in D_t$}{                 
                        $X,y \leftarrow \text{mini-batch of samples}$ \\
                        $\hat{y}, \ell^{str.} \leftarrow  \text{Forward Model}(\mathcal{F}_{\theta,\phi}, X, z^{\prime}, t)$ \\
                        \uIf{\text{new module added in last $k$ epochs}}{ 
                            \tcp{projection phase}
                            $\ell = \ell^{(str.)}$\\
                             \uIf{\text{use functional loss in projection}}{ 
                                $\ell += \mathcal{L}^{(fnc.)}(\hat{y}, y)$
                             }
                        }\uElse{
                            $\ell = \mathcal{L}^{(fnc.)}(\hat{y}, y)+\ell^{(str.)}$ \\
                        }
                    Update parameters $\theta, \phi$ using $\nabla_{\theta,\phi} \ell$
                    }
                } 
                \ForEach{module $m \in \mathcal{F}$}{
                    \text{Fix structural parameter $\phi$ of module $m$}
              }
            }
            \caption{Modular training with projection}
        \label{algo:lmc}
        \end{algorithm2e}
    \end{minipage}
    \hspace{1em}
    \begin{minipage}{0.55\linewidth}
        \setlength{\textfloatsep}{3mm}
        \begin{algorithm2e}[H]
        \small
        \SetAlgoLined
        \DontPrintSemicolon    
        \textbf{Require:}  Learner $\mathcal{F}_{\theta,\phi}$ with $L$ layers, batch $X$, $z$-score threshold $z^{\prime}$, task index $t$\\
        \textbf{Output:}  logits $\hat{y}$, structural loss $\ell^{(st.)}$ \\
        $x^{(0)}=X$\\
        $\ell^{(str.)}=0$\\
        \For{$l \in L$}{
                  Let $M^{(l)}$ denote a set of modules at layer $l$\\
			      Calculate:\\
  			      $\gamma^{(l)}\in \mathbb{R}^{|M^{(l)}|\times|X|}$ using Eq.~\ref{eq:gamma_l},\\
			      $w^{(l)}\in \mathbb{R}^{|M^{(l)}|\times|X|}$ using Eq.~\ref{eq:w_l_batch} or~\ref{eq:w_l},\\
			      $z^{(l)}\in \mathbb{R}^{|M^{(l)}|\times|X|}$ using Eq.~\ref{eq:z_score}, average over $|X|$\\
			      $x^{(l)}$ using Eq.~\ref{eq:x_l} \\
			      
                \uIf{$t>0$}{
                    \ForEach{$m\in M^{(l)}$}{      
                        \uIf{$z^{(l)}_m > z^\prime$}{ 
                            \uIf{\text{module was added at layer $l$ during task $t$}}{ 
                                $w^{(l)}_m=\vec{0}$ \tcp{does not use outlier modules}
                                }
                            \uElseIf{\text{\textbf{no} module added in last $k$ epochs}}{ 
                                \tcp{not in the projection phase}
                                Fix all modules at layer $l$ \\
                                Add a new free module to layer $l$\\
                            }
                        }
                     }
                }
                
            $\ell^{(str.)} = \ell^{(str.)} - \sum_m \text{\textit{sum}}\{\vec{\gamma}^{(l)}_m \odot \vec{w}^{(l)}_m\} $\\
			} 
			$\hat{y}=x^{(L)}$ \\
			\Return $\hat{y}, \ell^{(str.)}$
            \caption{Forward Model}
        \label{algo:forward}
        \end{algorithm2e}
    \end{minipage}
    }
}

\subsection{Structural component}\label{app:structural_component}
In practice, we applied the $log$-operation to the structural loss for both choices of the structural component, which resulted in a more stable training procedure.
\subsubsection{Invertible architectures}                                              
Invertible architectures, such as the one proposed by \citet{dinh2014nice}, can be used to model high-dimensional density after mapping the data in a space with some desirable factorization properties. We use this idea here to directly approximation of the activation likelihood of a module $m$. More specifically, \cite{hocquet2020ova} show that maximizing the likelihood of a module under such invertible transformation is equivalent to minimizing the $L_2$ norm of the output of structural component $s(o; \phi_m)$, yielding the local structural objective:
\begin{equation}
    \mathcal{L}^{(str.)}_m(x) = ||x||^2.
\end{equation}
To satisfy the invertibility constraint \cite{dinh2014nice} propose to split the input $o$ into blocks of equal size $o_1$ and $o_2$ and apply two, not necessarily invertible, transformations $s_1$ and $s_2$ as:
\begin{equation}
\label{eq:structural_output}
    \begin{split}
   a_1 = s_1(o_2;\phi_{2,m}) + o_1, \\
   a_2 = s_2(a_1;\phi_{1,m}) + o_2.
    \end{split}
\end{equation}                          

The output of structural component $a$ is obtain through the concatenation of $a_1$ and $a_2$. Importantly, the input and output of the structural component have the same dimensionality $a_m, o_m \in \mathbb{R}^k$. The inverse can be obtained as:
\begin{equation}
    \label{eq:s_1_s_2}
    \begin{split}
   o_2 = a_2 - s_2(a_1;\phi_{2,m}), \\
   o_1 = a_1 - s_1(o_2;\phi_{1,m}).
    \end{split}
\end{equation}         
Intuitively, the invertibility constraint prevents $a_1$ and $a_2$ from collapsing to the solution of outputting $0$-vectors, which would be useless.
\subsubsection{Other possible choices of structural component}
The role of structural component in LMC is to detect in-distribution and out-of-distribution samples for each module. It is natural to consider density estimates produced by deep generative models (DGM) for this task. In this work we only considered a simple encoder-decoder based architecture and a simple flow-model. Applying other more complex DGMs such as VAEs~\citep{kingma2013auto} or flow-based~\citep{behrmann2019invertible,kobyzev2020normalizing} models might further improve the efficacy of local structural component. Nevertheless, such models also come with their challenges, which include calibration difficulties as well as low data efficiency~\citep{wang2020further}.

\subsection{Architecture details}\label{app:architecture}
Unless otherwise stated, we initialize the learner with a single module per layer, each learned consist of 4 layers. The used architecture of each module is detailed in Table~\ref{tab:architectures}. For the CTrL experiments we used invertible structural component for the task-specific output heads (classifiers) in task ID agnostic LMC(A), while for feature extracting trunk we used an encoder-decoder architecture (i.e. structural component tasked to reconstruct the module's input).

Unless stated otherwise, architectures used for all baselines closely resemble the architecture used by LMC. Thus, by default all baselines contain 4 layers with each layer's architecture and parameter count being equivalent to the architecture and the parameter count of the functional component of an LMC module. In modular baselines (MNTDP, SG-F) each module corresponds to the functional component of the LMC's module. Some of the fixed capacity baselines in Table~\ref{tab:tab1} (e.g. HAT, EWC, Finetune L) where initialized with the layer width scaled up to match the parameter count of the largest possible modular network (e.g. in case of linear growth with one new module per layer per task the largest possible modular network in our framework would have 24 modules in a 6 tasks sequence).

\begin{table*}[!htb]         
    \caption{Used architectures per module: column 'layer $\mathcal{F}$' refers to the index of the layer in the modular learner $\mathcal{F}$ (i.e. 0 is the closes layer to the input), while 'layer m' gives the layer index within the module. Note, both linear layers in (c) are used in parallel as proposed in \citep{dinh2014nice} (i.e. $s_1$ and $s_2$ from Eq.~\ref{eq:s_1_s_2}).}

\begin{minipage}{1\linewidth}
    \begin{minipage}{.5\linewidth}
      \centering
    \setlength{\tabcolsep}{0.8pt}
    
  \begin{scriptsize}
    \begin{tabular}{l|c|c|ccccc}
    \toprule
     \multicolumn{8}{c}{Functional component} \\
    \midrule
     Type & layer $\mathcal{F}$ & layer m & \#params. & \#out ch. & stride & padding & kernel \\
    \midrule
     Conv. &  0 & 0 & 1792 & 64 & 1 & 2 & 3\\
     Conv. & 1-3 & 0 & 36928 & 64 & 1 & 2 & 3\\
     Batch norm & all & 1 & 128 & - & - & - \\
     ReLu & all & 2 & - & - & - & - \\
     Max. Pool & all & 3 & - & - & - & 2 & 2 \\
    \toprule
    \end{tabular}
  \end{scriptsize}
   \end{minipage}%
    \begin{minipage}{.5\linewidth}
      \centering
  \setlength{\tabcolsep}{0.8pt}
  \begin{scriptsize}
    \begin{tabular}{l|c|c|ccccc}
    \toprule
     \multicolumn{8}{c}{Structural component (decoder)} \\
    \midrule
     Type & layer $\mathcal{F}$ & layer m &\#params. & \#out ch. & stride & padding & kernel \\
    \midrule
     ConvTranspose2d &  all & 0 & 16448 & 64 & 2 & 2 & 2 \\
     Batch norm & all & 1 & 128 & - & - & - \\
     ReLu & all & 2 & - & - & - & - \\
     Conv. &  all & 3 & 65600 & 3 & 1 & 1 & 3\\
     Sigmoid & all & 4 & - & - & - & - \\
    \toprule
    \end{tabular}
  \end{scriptsize}
    \end{minipage} 
\end{minipage}
\begin{minipage}{1\linewidth}
  \centering
  \begin{minipage}{.5\linewidth}
  \centering
  \setlength{\tabcolsep}{0.8pt}
  \begin{scriptsize}
    \begin{tabular}{l|c|c|cccc}
    \toprule    
     \multicolumn{5}{c}{Structural component (invertible)} \\
    \midrule
     Type & layer $\mathcal{F}$ & layer m & \#params. & \#input & \#output \\
    \midrule
     Input $L_2$ Norm. & output head & 0 & - & - \\ 
     Linear  & output head & 1 & 83232 & 288 & 288  \\
     Linear  & output head & 1 & 83232 & 288 & 288  \\
    \toprule
    \end{tabular}
  \label{tab:architectures}
  \end{scriptsize}
\end{minipage} 
\end{minipage}
    
\end{table*} 

\subsection{Dealing with batch-norms and data normalisation}                   
While batch normalization~\citep{ioffe2015batch} is a useful device for accelerating the training of neural networks, it comes with challenges when it comes to settings with shifting data distribution such as meta-~\citep{antoniou2018train} and continual learning. Specifically, in continual learning when testing on the previously seen tasks after new tasks have been learned, the batch norm will change its statistics resulting in forgetting even if the parameters of the network have not been changed~\cite{gupta2020unreasonable}. We highlight several ways to deal with it. One way is to warm-up batch norm before testing on previous tasks by performing several forward passes through the model with unlabeled test data to let batch norms ``relearn'' the task statistics. This assumes that the test data is available in high quantity at test time (i.e. we cannot warm-up batch-norm with a single test sample).  Another way is to fix batch norms completely after a task has been learned. In monolithic architectures, such fixing might limit the plasticity of the network and prevent the learning of new tasks. In modular architectures, however, batch norms of frozen modules can be kept frozen while new module's batch norms can keep learning resulting in a balance between stability and plasticity. In monolithic architecture which are task-ID aware at test time, a separate batch-norm layer can be used per task.

In modular methods we fixed the batch norms whenever the modules are fixed (e.g. after learning a task in LMC and MNTDP). In HAT we used a separate batch-norm per task, as using a single batch-norm resulted in high forgetting rates. For other non-modular methods (e.g. EWC, ER) we used a single batch-norm layer for all tasks. Additionally, we observed that monolithic methods that share batch-norms across tasks result in high forgetting rate if no data normalisation is performed (and no batch-norm warm-up before testing). In the CTrL experiments we normalised tasks' data for all methods but LMC using statistics calculated on each task separately. In these experiments LMC performed better on not normalised data. In the OOD generalization experiment on cMNIST we normalised data also for LMC. Additionally, we performed batch-norm warm before testing in the OOD experiments. In meta-CL (\S~\ref{app:meta-cl}) the batch norms do not keep the estimates of running statistics ($track\_running\_stats$ is set to False) and momentum is set to 1. 
\bgroup

\section{Continual Transfer Learning (CTrL)}\label{app:ctrl}
\subsection{Streams used}
As in \cite{veniat2020efficient}, all input samples were reshaped to a 32 x 32 pixels resolution. We normalized for all methods but LMC unless otherwise stated. We did not use any data augmentation techniques. The datasets used for the first 5 streams are described in the Table~\ref{tab:split_details}, datasets used for the streams $S^{long30}$ and $S^{long}$ are described in the Table~\ref{tab:split_details_long_1} and Table~\ref{tab:split_details_long_2}.
\label{app:ctrl_streams}
\begin{table*}[ht]
  \centering
  \centerline{    
  \begin{scriptsize}
  \begin{sc}
  \renewcommand{\arraystretch}{0.7}
    \begin{tabular}{l|lcccccc}
    \toprule
    Stream & & $T_1$ & $T_2$ & $T_3$ & $T_4$ & $T_5$ & $T_6$  \\
    \midrule
    \multirow{3}{0.5cm}{$S^+$} & \textbf{Datasets}& Cifar-10\citep{krizhevsky2009learning} & MNIST\citep{lecun-mnisthandwrittendigit-2010} & DTD\citep{cimpoi14describing} & F-MNIST\citep{xiao2017fashion} & SVHN\citep{netzer2011reading} & Cifar-10 \\
     & \textbf{$\#$Train samples}  & 4000 & 400 & 400 & 400 & 400 & 400  \\
     & \textbf{$\#$Val. samples}  & 2000 & 200 & 200 & 200 & 200 & 200  \\
    \multirow{3}{0.5cm}{$S^-$} & \textbf{Datasets} & Cifar-10 & MNIST & DTD & F-MNIST & SVHN & CIFAR-10  \\
     & \textbf{$\#$Train samples}  & 400 & 400 & 400 & 400 & 400 & 4000 \\
     & \textbf{$\#$Val. samples}  & 200 & 200 & 200 & 200 & 200 &  2000 \\
    \multirow{3}{0.5cm}{$S^{in}$} & \textbf{Datasets} & R-MNIST  & Cifar-10 & DTD  & F-MNIST  & SVHN & R-MNIST \\
     & \textbf{$\#$Train samples}  & 4000 & 400 & 400 & 400 & 400 &  50\\
     & \textbf{$\#$Val. samples}  & 2000 & 200 & 200 & 200 & 200 & 30 \\
    \multirow{3}{0.5cm}{$S^{out}$} & \textbf{Datasets} &  CIFAR-10 & MNIST & DTD & F-MNIST & SVHN & Cifar-10 \\
    & \textbf{$\#$Train samples}  & 4000 & 400 & 400 & 400 & 400 & 400  \\
     & \textbf{$\#$Val. samples}  & 2000 & 200 & 200 & 200 & 200 & 200  \\
    \multirow{3}{0.5cm}{$S^{pl}$} & \textbf{Datasets} & MNIST & DTD & F-MNIST & SVHN &  Cifar-10 & \\
     & \textbf{$\#$Train samples}  & 400 & 400 & 400 & 400 & 4000 &   \\
     & \textbf{$\#$Val. samples}  & 200 & 200 & 200 & 200 & 2000 & \\
     \toprule
    \end{tabular}
    \end{sc}
    \end{scriptsize}
    \caption{Details on the datasets and training/validation data amounts for the used streams~\citep{veniat2020efficient}.}
    \label{tab:split_details}
    }
\end{table*}
\egroup

\begin{table*}[ht]
  \centering
  \centerline{    
  \begin{scriptsize}
  \begin{sc}
  \renewcommand{\arraystretch}{0.7}
    \begin{tabular}{l|ccccc}
    \toprule
    Task & Dataset & Classes & \# Train & \# Val & \# Test \\
    \midrule
    0 & cifar10  & deer, truck, dog, cat, bird&25 & 15 & 5000  \\ 
1 &  mnist  & mnist 6 - six,0 - zero,7 - seven,8 - eight,4 - four&5000 & 2500 & 4894  \\ 
2 & fashion-mnist &  Coat,  Bag,  Trouser,  Dress,  T-shirt/top&5000 & 2500 & 5000  \\ 
3 &  svhn  & svhn 9 - nine,8 - eight,4 - four,0 - zero,6 - six&25 & 15 & 5000  \\ 
4 & cifar100  & worm, possum, aquarium fish, orchid, lizard&25 & 15 & 500  \\ 
5 & cifar10  & frog, automobile, cat, truck, dog&5000 & 2500 & 5000  \\ 
6 &  svhn  & svhn 3 - three,1 - one,5 - five,4 - four,7 - seven&25 & 15 & 5000  \\ 
7 &  mnist  & mnist 4 - four,5 - five,3 - three,2 - two,7 - seven&5000 & 2500 & 4874  \\ 
8 & fashion-mnist &  Sneaker,  Sandal,  Ankle boot,  Coat,  T-shirt/top&25 & 15 & 5000  \\ 
9 & fashion-mnist &  Dress,  Coat,  Ankle boot,  Bag,  Trouser&5000 & 2500 & 5000  \\ 
10 &  svhn  & svhn 3 - three,7 - seven,0 - zero,1 - one,8 - eight&25 & 15 & 5000  \\ 
11 & cifar100  & otter, leopard, beetle, ray, butterfly&2250 & 1250 & 500  \\ 
12 &  svhn  & svhn 6 - six,1 - one,9 - nine,2 - two,0 - zero&25 & 15 & 5000  \\ 
13 & fashion-mnist &  Sneaker,  Ankle boot,  T-shirt/top,  Sandal,  Dress&5000 & 2500 & 5000  \\ 
14 &  mnist  & mnist 5 - five,1 - one,9 - nine,7 - seven,8 - eight&5000 & 2500 & 4866  \\ 
15 &  mnist  & mnist 5 - five,6 - six,7 - seven,9 - nine,2 - two&25 & 15 & 4850  \\ 
16 &  svhn  & svhn 4 - four,0 - zero,1 - one,2 - two,7 - seven&25 & 15 & 5000  \\ 
17 & fashion-mnist &  T-shirt/top,  Sneaker,  Shirt,  Trouser,  Sandal&25 & 15 & 5000  \\ 
18 & cifar10  & cat, frog, bird, ship, deer&5000 & 2500 & 5000  \\ 
19 &  svhn  & svhn 9 - nine,2 - two,8 - eight,4 - four,7 - seven&25 & 15 & 5000  \\ 
20 & cifar10  & ship, horse, dog, truck, cat&25 & 15 & 5000  \\ 
21 & fashion-mnist &  Sneaker,  T-shirt/top,  Shirt,  Dress,  Pullover&5000 & 2500 & 5000  \\ 
22 & cifar10  & airplane, truck, deer, frog, bird&5000 & 2500 & 5000  \\ 
23 &  svhn  & svhn 2 - two,6 - six,4 - four,1 - one,5 - five&5000 & 2500 & 5000  \\ 
24 &  mnist  & mnist 8 - eight,3 - three,9 - nine,4 - four,7 - seven&25 & 15 & 4956  \\ 
25 &  svhn  & svhn 4 - four,8 - eight,2 - two,6 - six,7 - seven&25 & 15 & 5000  \\ 
26 &  svhn  & svhn 1 - one,4 - four,7 - seven,9 - nine,2 - two&25 & 15 & 5000  \\ 
27 & cifar100  & sweet pepper, cockroach, motorcycle, tank, elephant&25 & 15 & 500  \\ 
28 &  svhn  & svhn 3 - three,2 - two,4 - four,7 - seven,1 - one&5000 & 2500 & 5000  \\ 
29 & cifar100  & chimpanzee, streetcar, wolf, beaver, rose&25 & 15 & 500  \\ 
30 & cifar10  & horse, airplane, deer, automobile, truck&25 & 15 & 5000  \\ 
31 &  svhn  & svhn 5 - five,8 - eight,7 - seven,4 - four,3 - three&5000 & 2500 & 5000  \\ 
32 & fashion-mnist &  Coat,  Dress,  Sandal,  Pullover,  T-shirt/top&5000 & 2500 & 5000  \\ 
33 & cifar10  & horse, ship, truck, frog, cat&25 & 15 & 5000  \\ 
34 & cifar10  & ship, dog, bird, airplane, cat&25 & 15 & 5000  \\ 
35 & cifar10  & deer, airplane, ship, truck, automobile&5000 & 2500 & 5000  \\ 
36 & cifar100  & boy, beaver, willow tree, shark, tank&25 & 15 & 500  \\ 
37 &  svhn  & svhn 3 - three,4 - four,9 - nine,1 - one,8 - eight&25 & 15 & 5000  \\ 
38 &  svhn  & svhn 9 - nine,4 - four,5 - five,3 - three,1 - one&25 & 15 & 5000  \\ 
39 & cifar10  & frog, airplane, cat, dog, truck&25 & 15 & 5000  \\ 
40 & cifar10  & ship, deer, truck, horse, bird&25 & 15 & 5000  \\ 
41 & fashion-mnist &  Dress,  Shirt,  Trouser,  Coat,  Sneaker&25 & 15 & 5000  \\ 
42 & cifar100  & streetcar, beaver, tiger, bus, raccoon&25 & 15 & 500  \\ 
43 & fashion-mnist &  Coat,  Bag,  Dress,  Sneaker,  Sandal&25 & 15 & 5000  \\ 
44 &  mnist  & mnist 5 - five,9 - nine,7 - seven,6 - six,2 - two&5000 & 2500 & 4850  \\ 
45 & cifar100  & hamster, pine tree, cockroach, boy, couch&25 & 15 & 500  \\ 
46 &  mnist  & mnist 0 - zero,3 - three,2 - two,7 - seven,9 - nine&5000 & 2500 & 4980  \\ 
47 & fashion-mnist &  Sandal,  Dress,  Coat,  Trouser,  Bag&25 & 15 & 5000  \\ 
48 &  svhn  & svhn 0 - zero,8 - eight,5 - five,2 - two,1 - one&5000 & 2500 & 5000  \\ 
49 & cifar10  & horse, frog, airplane, dog, ship&5000 & 2500 & 5000  \\ 
    \toprule
    \end{tabular}
    \end{sc}
    \end{scriptsize}               
    \caption{Details on the datasets and training/validation data amounts used for $S^{long}$ (Part 1)~\citep{veniat2020efficient}.}
    \label{tab:split_details_long_1}
    }
\end{table*}

\begin{table*}[ht]
  \centering
  \centerline{    
  \begin{scriptsize}
  \begin{sc}
  \renewcommand{\arraystretch}{0.7}
    \begin{tabular}{l|ccccc}
    \toprule
    Task & Dataset & Classes & \# Train & \# Val & \# Test \\
    \midrule
    50 &  svhn  & svhn 9 - nine,4 - four,6 - six,5 - five,2 - two&25 & 15 & 5000  \\ 
51 &  svhn  & svhn 3 - three,6 - six,8 - eight,9 - nine,1 - one&25 & 15 & 5000  \\ 
52 & cifar100  & crocodile, lion, butterfly, otter, hamster&2250 & 1250 & 500  \\ 
53 &  mnist  & mnist 9 - nine,8 - eight,6 - six,7 - seven,3 - three&25 & 15 & 4932  \\ 
54 &  mnist  & mnist 7 - seven,3 - three,8 - eight,4 - four,2 - two&25 & 15 & 4956  \\ 
55 &  svhn  & svhn 4 - four,2 - two,6 - six,0 - zero,5 - five&25 & 15 & 5000  \\ 
56 & cifar100  & sea, chair, snake, spider, snail&25 & 15 & 500  \\ 
57 & cifar100  & beetle, television, table, porcupine, cup&25 & 15 & 500  \\ 
58 & cifar10  & cat, horse, frog, truck, automobile&25 & 15 & 5000  \\ 
59 &  svhn  & svhn 8 - eight,6 - six,1 - one,5 - five,3 - three&25 & 15 & 5000  \\ 
60 & cifar10  & bird, frog, horse, ship, deer&25 & 15 & 5000  \\ 
61 &  mnist  & mnist 1 - one,9 - nine,8 - eight,7 - seven,2 - two&25 & 15 & 4974  \\ 
62 & fashion-mnist &  Dress,  T-shirt/top,  Sandal,  Trouser,  Sneaker&25 & 15 & 5000  \\ 
63 &  mnist  & mnist 6 - six,4 - four,0 - zero,7 - seven,8 - eight&25 & 15 & 4894  \\ 
64 &  svhn  & svhn 4 - four,2 - two,7 - seven,6 - six,3 - three&5000 & 2500 & 5000  \\ 
65 & cifar100  & pear, skyscraper, shark, plain, dolphin&2250 & 1250 & 500  \\ 
66 & cifar10  & frog, bird, airplane, ship, horse&25 & 15 & 5000  \\ 
67 & cifar10  & frog, deer, ship, horse, truck&25 & 15 & 5000  \\ 
68 & cifar10  & horse, deer, truck, airplane, dog&25 & 15 & 5000  \\ 
69 & cifar100  & skunk, orchid, cattle, spider, lobster&25 & 15 & 500  \\ 
70 &  mnist  & mnist 3 - three,5 - five,4 - four,9 - nine,1 - one&25 & 15 & 4874  \\ 
71 &  svhn  & svhn 4 - four,3 - three,1 - one,7 - seven,5 - five&25 & 15 & 5000  \\ 
72 & fashion-mnist &  Coat,  Dress,  Bag,  Sandal,  Trouser&25 & 15 & 5000  \\ 
73 & fashion-mnist &  Sandal,  Dress,  Ankle boot,  Pullover,  Shirt&25 & 15 & 5000  \\ 
74 &  mnist  & mnist 3 - three,2 - two,8 - eight,6 - six,4 - four&25 & 15 & 4914  \\ 
75 & cifar10  & airplane, dog, horse, bird, ship&25 & 15 & 5000  \\ 
76 & cifar10  & automobile, horse, airplane, cat, truck&25 & 15 & 5000  \\ 
77 & fashion-mnist &  Sandal,  Coat,  Shirt,  Dress,  Ankle boot&25 & 15 & 5000  \\ 
78 & fashion-mnist &  Trouser,  T-shirt/top,  Sandal,  Sneaker,  Dress&25 & 15 & 5000  \\ 
79 & cifar100  & lion, turtle, cup, shrew, rose&25 & 15 & 500  \\ 
80 &  mnist  & mnist 2 - two,4 - four,5 - five,6 - six,1 - one&25 & 15 & 4832  \\ 
81 & cifar100  & turtle, mountain, kangaroo, lobster, crab&25 & 15 & 500  \\ 
82 & fashion-mnist &  Sandal,  Sneaker,  T-shirt/top,  Coat,  Pullover&25 & 15 & 5000  \\ 
83 & cifar100  & plain, skyscraper, butterfly, train, sea&25 & 15 & 500  \\ 
84 &  mnist  & mnist 9 - nine,5 - five,4 - four,8 - eight,2 - two&25 & 15 & 4848  \\ 
85 &  svhn  & svhn 1 - one,7 - seven,0 - zero,5 - five,6 - six&25 & 15 & 5000  \\ 
86 &  mnist  & mnist 2 - two,4 - four,7 - seven,3 - three,8 - eight&25 & 15 & 4956  \\ 
87 & cifar10  & ship, automobile, frog, dog, horse&25 & 15 & 5000  \\ 
88 & cifar100  & cloud, spider, tiger, mouse, snake&25 & 15 & 500  \\ 
89 & fashion-mnist &  Dress,  Pullover,  T-shirt/top,  Bag,  Shirt&25 & 15 & 5000  \\ 
90 & cifar10  & automobile, truck, cat, dog, horse&25 & 15 & 5000  \\ 
91 &  mnist  & mnist 0 - zero,8 - eight,9 - nine,7 - seven,5 - five&25 & 15 & 4846  \\ 
92 &  mnist  & mnist 3 - three,9 - nine,7 - seven,5 - five,8 - eight&25 & 15 & 4866  \\ 
93 & fashion-mnist &  Bag,  Coat,  T-shirt/top,  Ankle boot,  Trouser&25 & 15 & 5000  \\ 
94 & cifar100  & camel, tractor, orchid, pear, aquarium fish&25 & 15 & 500  \\ 
95 &  mnist  & mnist 2 - two,8 - eight,9 - nine,4 - four,3 - three&25 & 15 & 4956  \\ 
96 &  mnist  & mnist 9 - nine,8 - eight,4 - four,0 - zero,7 - seven&25 & 15 & 4936  \\ 
97 & fashion-mnist &  Bag,  Dress,  Shirt,  Sandal,  Pullover&25 & 15 & 5000  \\ 
98 & cifar100  & mouse, snail, bed, trout, girl&25 & 15 & 500  \\ 
99 & fashion-mnist &  Trouser,  Pullover,  Sandal,  T-shirt/top,  Ankle boot&25 & 15 & 5000  \\ 

    \toprule
    \end{tabular}
    \end{sc}
    \end{scriptsize}               
    \caption{Details on the datasets and training/validation data amounts used for $S^{long}$ (Part 2).}
    \label{tab:split_details_long_2}
    }
\end{table*}

\subsection{Baselines and training details}\label{app:baseline_details}  
We adopted the original soft-gating with fixed modules (\textbf{SG-F}) proposed by \citet{mendez2020lifelong} in two ways: (1) instead or relying on a pool of initially pretrained modules shared across all layers, we initialize a separate set of modules per layer. This is necessary in order to comply with the experimental setup of CTrL which does not allow pretraining. (2) We used the expansion strategy proposed in MNTDP~\cite{veniat2020efficient} for SG-F, i.e. for each task different layouts with no or one new module per-layer starting at the top layer are trained, the layout with the best validation accuracy is accepted. The original expansion strategy of \citep{mendez2020lifelong} is similar in spirit, yet relies on a module pool shared between layers and an initial pretraining of modules, which allows training of only two parallel models: with and without adding a single new module to the shared pool.

In \textbf{SG-F\tiny{(A)}} we share a single controller network among all tasks in the sequence. Thereby, the main network still uses task IDs to select the task-specific output head. In the controller a single head architecture is used to gate the modules. As modules are added to the learner, each head of the controller used to gate each layer of the main learner is also expanded. This baseline showcases forgetting in the controller if it is shared across tasks.

For \textbf{HAT} we used a separate batch-norm layer for each task. Using shared batch-norm resulted in high forgetting rate for this method.

For all task ID aware methods, the task ID was used either to select the task specific output head (as in HAT, EWC or ER) or the task specific structure as in MNTDP (which includes the output head). Thereby, we treat first and last tasks in $S^+$, $S^-$, $S^{in}$ and $S^{out}$ streams as tasks with different IDs. This corresponds to the definition provided in \cite{veniat2020efficient}, where the task ID is defined to correspond to the sequential order of the task in the sequence.

We used Adam optimizer for all baseline methods but HAT~\citep{Serra18}. Using Adam for HAT resulted in more forgetting, which we believe is because HAT masks out only the gradients of some parameters and does not effect Adam's momentum.

\textbf{Hyper-parameter and model selection} was performed using average mean validation accuracy over all tasks in the stream (stream level) with splits detailed in Table~\ref{tab:split_details}.  When varying the seeds in the provided experiments, we did not very the seed that effects data generation (CTrL Streams) but only the seed that affected the algorithm, model initialization as well as data-loader's batch sampling.

\subsection{Metrics}
\label{app:ctrl_metrics} 
Here we formally define metrics used in the experiments. These metrics are similar to the ones used by \citet{veniat2020efficient}. $\Delta$ denotes the prediction accuracy of the predictor $\mathcal{F}$. We use subscripts to indicate the version of the parameters: e.g. $\theta_{1\ldots t}$ indicates the functional parameters of the learner after it was continually trained on $t$ tasks, while $\theta_T$ indicates the version of functional parameters after learning only task $T$ in isolation.
\paragraph{Average accuracy} on all tasks seen so far.
\begin{align}
    \mathcal{A} = \frac{1}{T} \sum_{t=1}^T \mathbb{E}_{(x,y)\sim\mathcal{D}_t}[\Delta\Big(\mathcal{F}(x;\theta_{1\ldots T},\phi_{1\ldots T}), y)\Big)] 
\end{align}
\paragraph{Forgetting}--- the average loss of accuracy on a task at the end of training as compared to the first time the task was seen. Positive values indicate positive backward transfer.
\begin{align}
\mathcal{F}=\frac{1}{T-1} \sum_{t=1}^T \mathbb{E}_{(x,y)\sim\mathcal{D}_t}[\Delta\Big(\mathcal{F}(x;\theta_{1\ldots T}, \phi_{1\ldots T}), y\Big) - \Delta\Big(\mathcal{F}(x; \theta_{1\ldots t}, \phi_{1\ldots t}), y\Big) ] 
\end{align}
\paragraph{Transfer} --- the difference in performance on the last ($T$'th) task between the modular learner trained on the entire sequence and an expert $\mathcal{F}^\prime$ trained on the last task in isolation.
\begin{align}
    \mathcal{T} = \mathbb{E}_{(x,y)\sim\mathcal{D}_T}\Delta\Big(\mathcal{F}(x; \theta_{1\ldots T}, \phi_{1\ldots T}), y\Big) - \Delta\Big(\mathcal{F}^\prime(x; \theta_T), y\Big)
\end{align}
 \newpage
\subsection{Transfer results CTrL}\label{app:transfer_ctrl}
\newcommand{\ttiny}[1]{{\fontsize{5}{5}\selectfont #1}}
\newcommand{\pmt}{\tiny\textpm}
\newcolumntype{g}{>{\columncolor{Gray}}c}
\newcolumntype{d}{>{\columncolor{Gray}}l}
\setlength{\tabcolsep}{0.5pt}

\begin{table*}[!h]
  \centering
  \centerline{    
  \begin{scriptsize}
  \begin{sc}             
  \renewcommand{\arraystretch}{1.15}
    \begin{tabular}{l|gcg|cgc|gcg|cgc}
    \toprule
    & \multicolumn{3}{c|}{$S^-$} & \multicolumn{3}{c|}{$S^+$} & \multicolumn{3}{c|}{$S^{in}$} &  \multicolumn{3}{c}{$S^{out}$} \\
    Model &  Acc $t_1$ & Acc $t^-_1$ & $\mathcal{T}$ &  Acc $t_1$ & Acc $t^+_1$ & $\mathcal{T}$ & Acc $t_1$ & Acc $t^{\prime}_1$ & $\mathcal{T}$ &  Acc $t_1$ & Acc $t^{\prime\prime}_1$ & $\mathcal{T}$\\
    \midrule
    Experts & 65.5\pmt0.7& 41.8\pmt1.0 & 0 & 41.3\pmt2.9 & 65.6\pmt0.5 & 0 & 98.5\pmt0.2 & 76.9\pmt4.9 & 0 & 65.9\pmt0.6 & 43.5\pmt1.6 & 0 \\
    MNTDP & 63.0\pmt3.6 & 56.9\pmt5.1 & 15.1 & 43.2\pmt0.7 & 65.9\pmt0.8 & 0.3 & 98.9\pmt0.1 & 93.3\pmt1.6 & 16.4 & 65.0\pmt1.2 & 57.7\pmt1.7 & 14.2  \\
    LMC\tiny{($\lnot$A)} & 65.2\pmt0.4 & 60.0\pmt1.1 & 18.2 & 42.9\pmt0.9 & 60.6\pmt1.9 & -4.7 & 98.7\pmt0.1 & 92.5\pmt7.6 & 15.6 & 65.2\pmt0.2 & 59.8\pmt1.1 & 16.3 \\
    LMC\tiny{(A)} & 65.2\pmt0.4 & 63.0\pmt1.7 & 21.2 & 43.1\pmt0.6 & 62.2\pmt0.7 & -3.4 & 98.7\pmt0.1 & 88.3\pmt1.6 & 11.4 & 65.5\pmt0.6 & 42.0\pmt21.9 & -1.5 \\
    S.G+FX & 64.9\pmt0.4 & 49.1\pmt7.3 & 7.3 & 43.1\pmt0.4 & 61.7\pmt1.7 & -3.9 & 98.8\pmt0.1 & 80.4\pmt6.8 & 3.5 & 65.0\pmt0.4 & 51.5\pmt6.5 & 8.0  \\
    \bottomrule
    \end{tabular}
  \end{sc}
  \end{scriptsize}
  }
  \caption{Transfer results on the CTrL benchmark. We provide the accuracy of the first and the last task on each of the streams. Additionally, we measure transfer $\mathcal{T}$ as the difference between the last task's accuracy of the model trained on the corresponding stream and an expert model trained on the last task in isolation.  }
 \label{tab:additional_transfer_results_ctrl}
\end{table*}

\subsection{Module selection: CTrL}
\begin{figure}[!h]%
    \subfloat[Module Selection $S^-$]{%
        \includegraphics[clip,width=1\columnwidth]{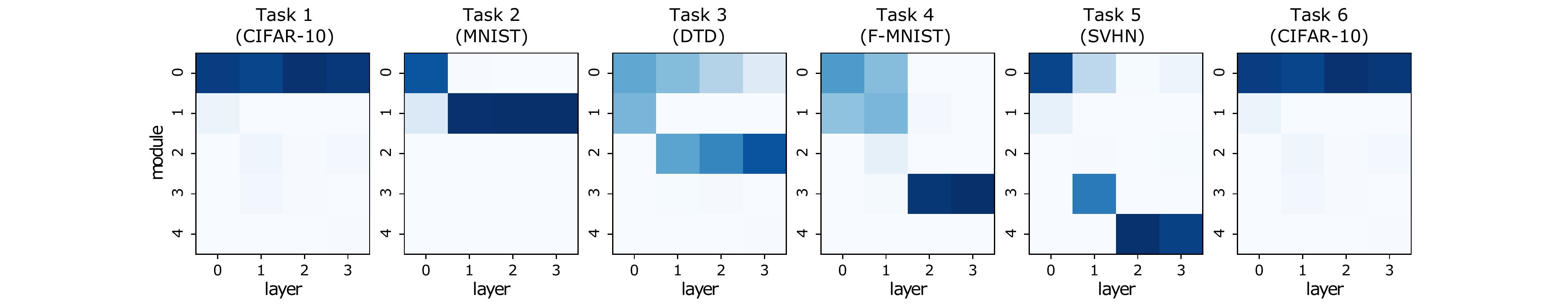}%
        }
    \vspace{0.5cm}
    \subfloat[Module Selection $S^+$]{%
      \includegraphics[clip,width=1\columnwidth]{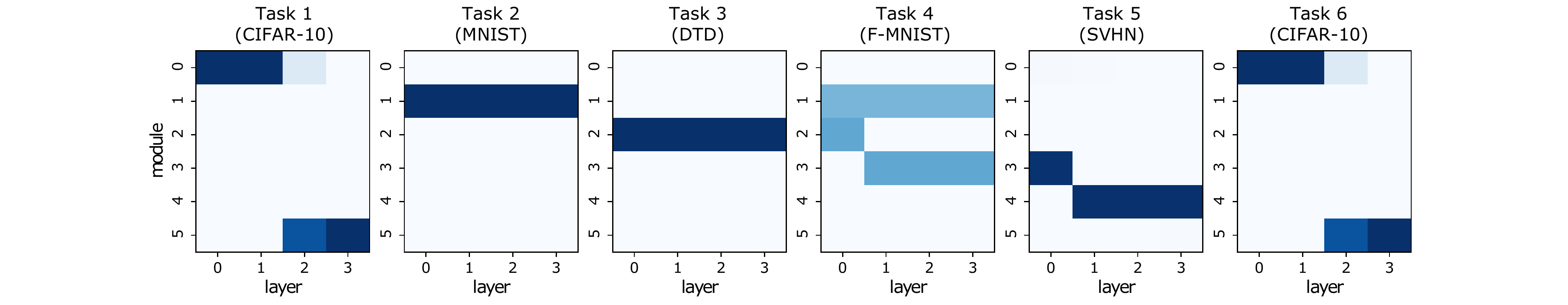}%
    }
    \vspace{0.5cm}           
    \subfloat[Module Selection $S^{in}$]{%
      \includegraphics[clip,width=1\columnwidth]{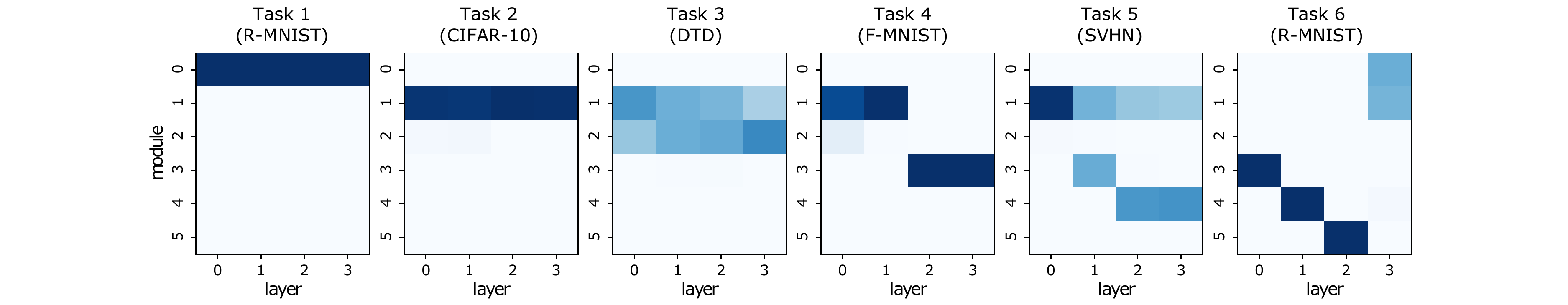}%
    }
    \vspace{0.5cm}
    \subfloat[Module Selection $S^{pl}$]{%
      \includegraphics[clip,width=1\columnwidth]{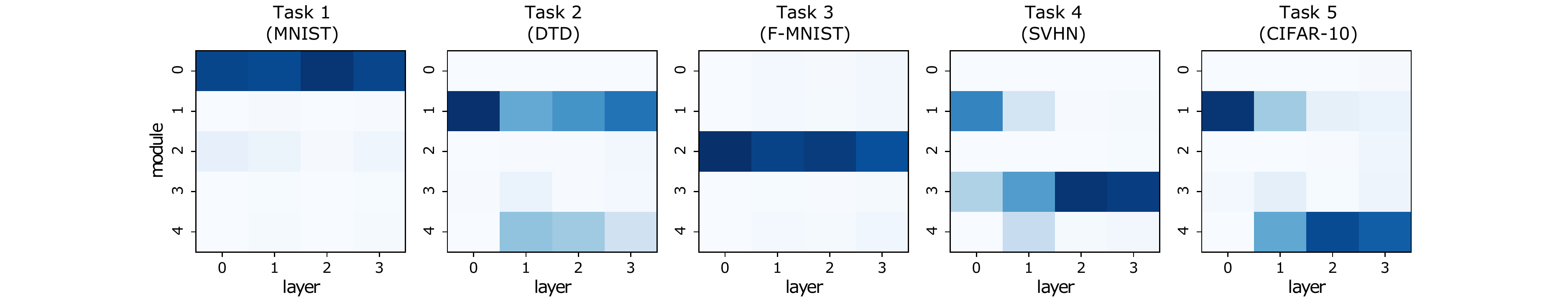}%
    }
    \caption{Average module selection for different streams at test time. Horizontal axis corresponds to the layer number, vertical axis corresponds to the module index at a layer; darker color corresponds to higher average activation strength (averaged over batches in the corresponding task).}
    \label{fig:module_selection_CTrL}
\end{figure}

In Figure~\ref{fig:module_selection_CTrL} we plot the average module selection of LMC for different streams of the CTrL~\citep{veniat2020efficient} benchmark after continual training on the corresponding stream. The plots correspond to the runs with the best validation accuracy on the corresponding stream. We observe that for the $S^{-}$ stream LMC reuses modules trained on the task 1 for the sixth task. On $S^{+}$, when tested on task 1 and task 6 same modules are used, yet the two modules at the last layer correspond to the ones added when learning task 6 (this task contains more training data than task 1). On $S^{in}$ the last task reuses only the last layer's module from task 1 and several other modules on the previous layers from previous tasks. Finally, for $S^{pl}$ stream, even though the tasks are unrelated, several modules are reused across the tasks. Importantly, since LMC implements soft module selections strategy several modules can be used together for the same batch of samples (see Eq.~\ref{eq:w_l}).

\subsection{Ablation of threshold $z^{\prime}$}       
\label{app:ablation_threshold}
\begin{figure}%
    \centering
    \subfloat[$\mathcal{A}$ and M for different $z^{\prime}$.]{{\includegraphics[width=0.45\textwidth]{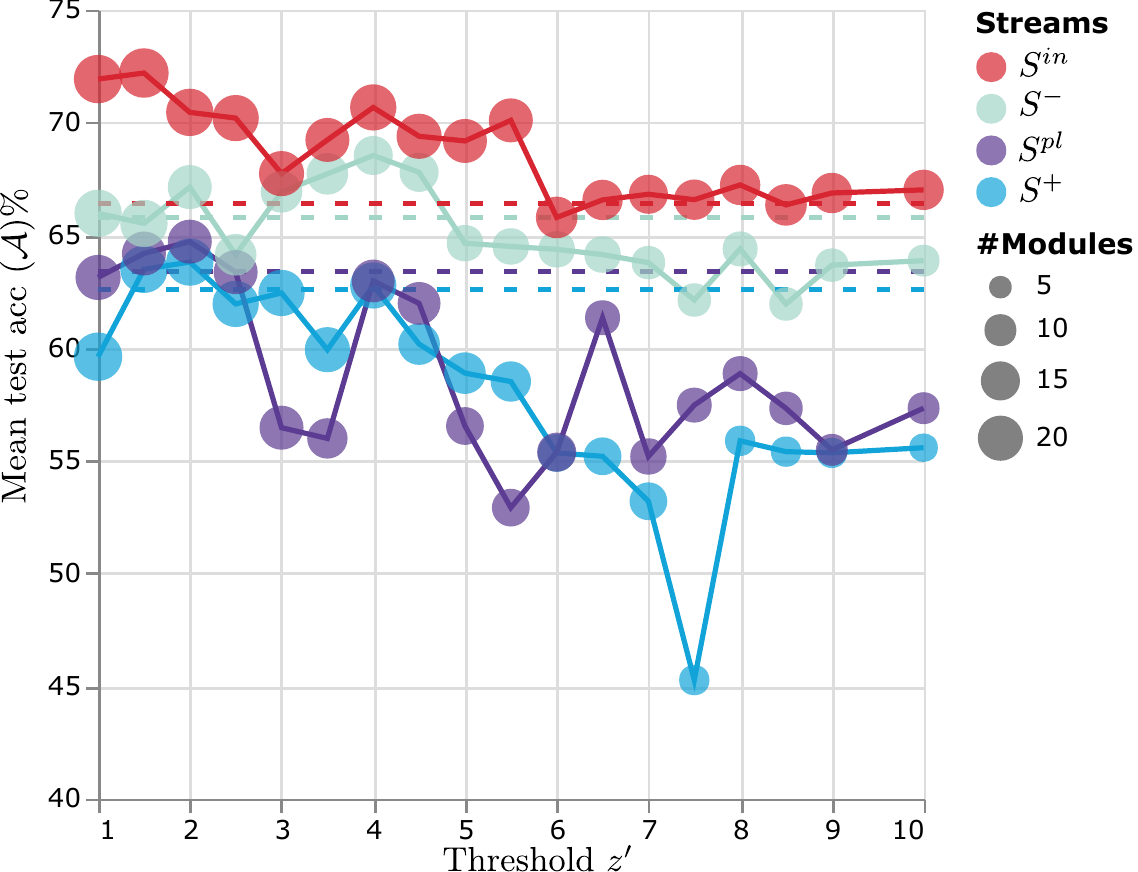} }
    \label{fig:acc_nmodules_threshold}
    }%
    \qquad       
    \subfloat[$\mathcal{A}$ for different M]{{\includegraphics[width=0.45\textwidth]{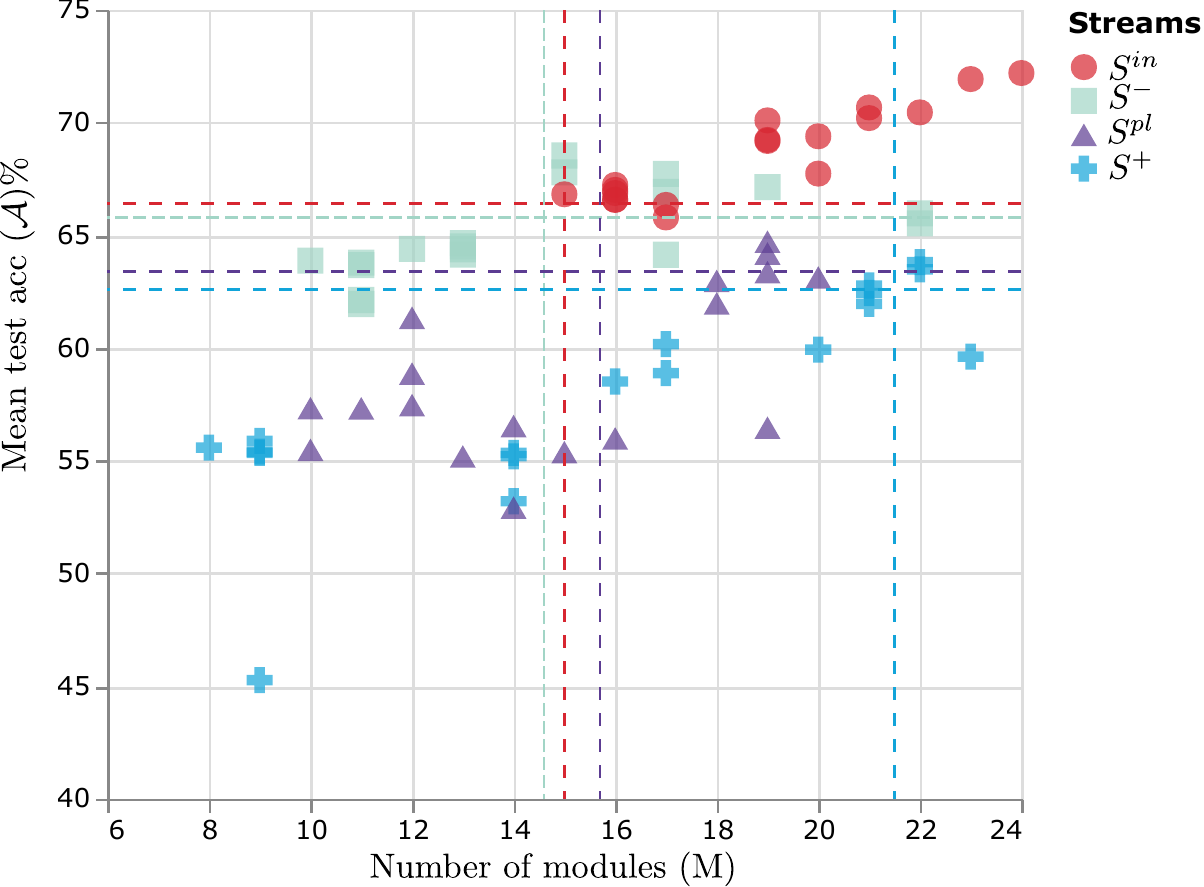}}
    \label{fig:numodules_acc}%
    }%
    \caption{(a) Mean test accuracy $\mathcal{A}$ and number of modules M for different values of the threshold hyperparameter $z^{\prime}$ for LMC{\tiny($\lnot$A)}. Dotted lines mark the accuracy of MNTDP\citep{veniat2020efficient}. (b) Number of modules (x-axis) against average accuracy $\mathcal{A}$ for runs with different threshold $z^{\prime}$. Dotted lines mark $\mathcal{A}$ and M for MNTDP\citep{veniat2020efficient} (best seen in color).}%
\end{figure}

To see how the z threshold influences the final number of modules (M) and the average (test) accuracy over seen tasks ($\mathcal{A}$) we report the $\mathcal{A}$ and M for LMC{\tiny($\lnot$A)} on 4 streams with fixed hyperparameters while only varying the threshold $z^{\prime}$. In Figure~\ref{fig:acc_nmodules_threshold} we plot the values of the threshold $z^{\prime}$ on the x-axis against average accuracy $\mathcal{A}$ on the y-axis: higher $z^{\prime}$ leads to fewer modules being instantiated resulting in lower average accuracy $\mathcal{A}$. Additionally, in Figure~\ref{fig:numodules_acc} we plot the same runs but now with number of modules on the x-axis and the average accuracy $\mathcal{A}$ on the y-axis. We identify the number of modules and accuracy of the MNTDP baseline with dotted lines with the corresponding stream colors in both plots. In  Figure~\ref{fig:numodules_acc} we observe that LMC instantiates a comparable number of modules as MNTDP in order to reach similar accuracy, with an exception of the $S^{pl.}$ stream, where LMC tends to add mode modules to reach a similar accuracy.

\section{Long sequences hyperparameter search visualization}
\label{app:long_modules_vs_acc}
\begin{figure}%
    \centering
    \subfloat[30 tasks sequence $S^{long30}$.]{{\includegraphics[width=0.6\textwidth]{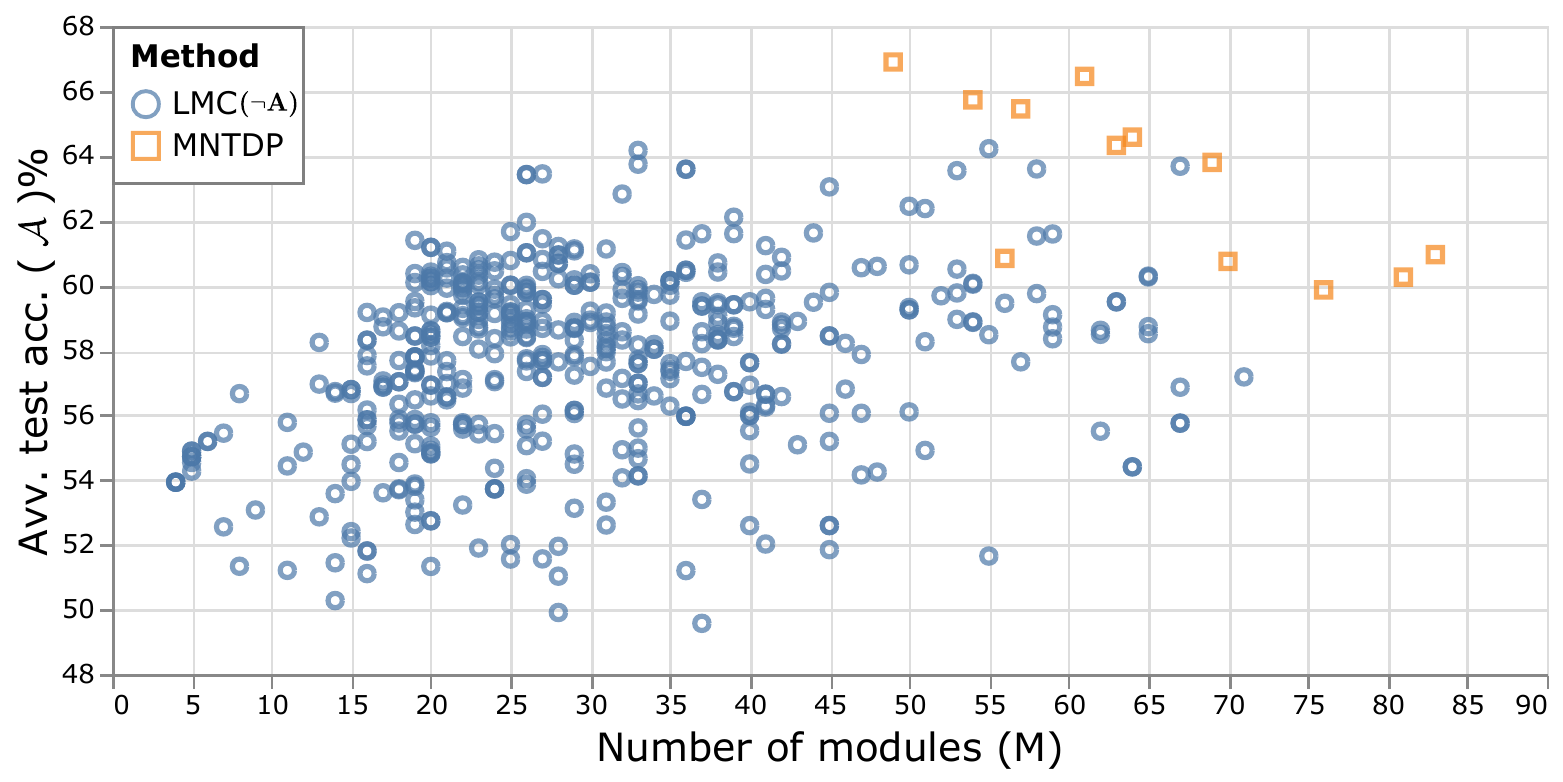} }
    }%
    \qquad   
    \subfloat[100 tasks sequence $S^{long}$.]{{\includegraphics[width=0.6\textwidth]{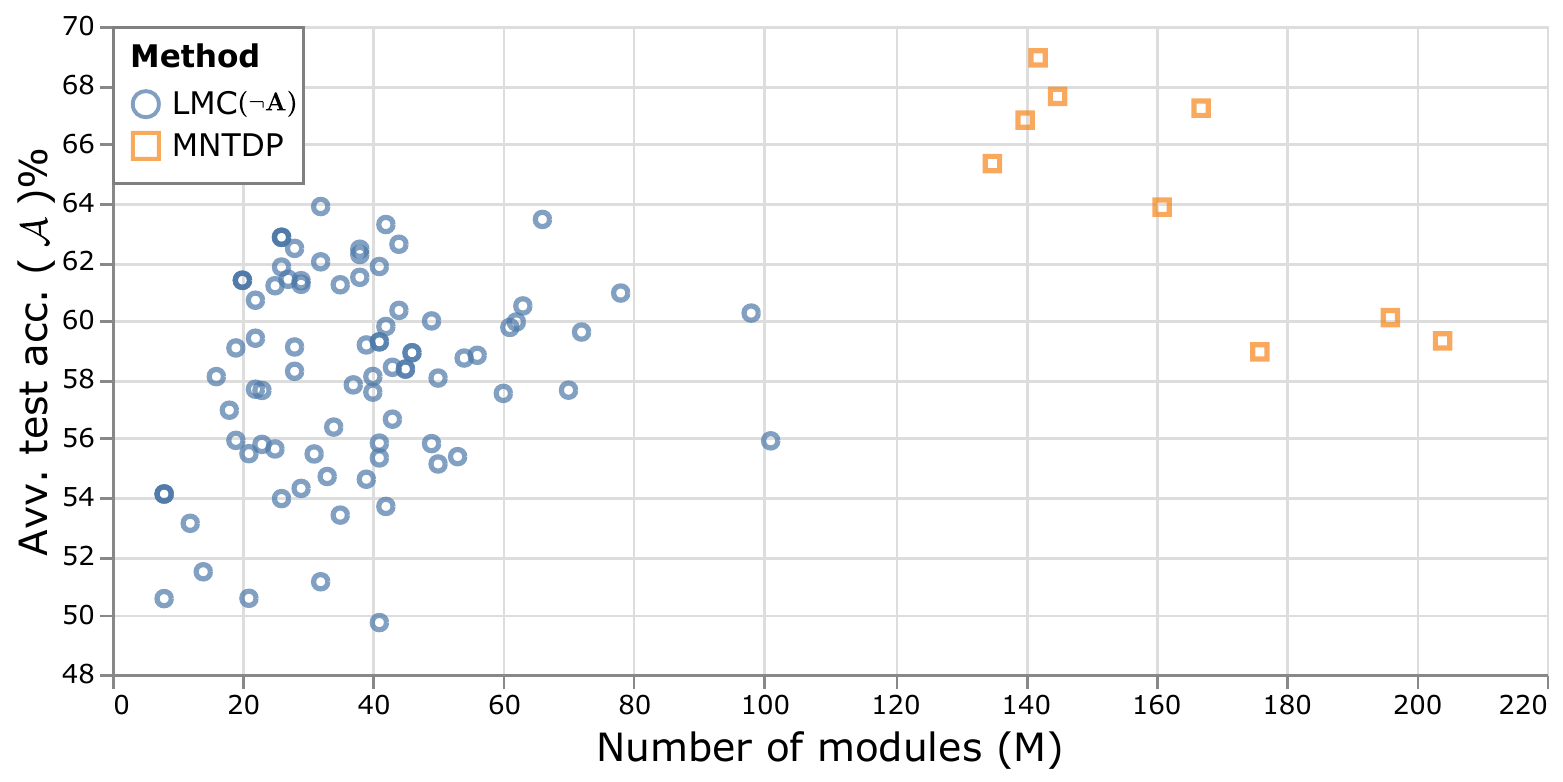}}}%
    \caption{Results on $S^{long}$ and $S^{long30}$ sequences for different hyperparameter values for all executed hyperparameter search runs (this is an expanded version of Figure~\ref{fig:acc_nmodules_slong}).}%
    \label{fig:long_modules_vs_acc}%
\end{figure}

We plot number of modules against the average test accuracy over all seen tasks ($\mathcal{A}$) in Figure~\ref{fig:long_modules_vs_acc} for all executed hyperparameter search runs on both $S^{long30}$ and $S^{long}$ sequences.
\section{Combining independent models}
\label{app:combining_independent_models}
The aim of this experiment is to show the ability of independently trained LMC models to be combined without fine-tuning, without loss in performance and also enabling positive transfer. In this experiment the test set's distributions of the cMNIST tasks are different from the training/validation sets' distributions (see \S~\ref{sec:pnp}). We used oracle model selection choosing hyperparameters on the test set. Note, that model selection is an unsolved challenge in the OOD generalization literature, where selecting model using oracle strategies is sometimes excepted if the baselines methods are also tuned using oracle strategies~\citep{gulrajani2020search,pezeshki2020gradient}.
\begin{figure*}[!h]%
    \centering                                          
    \includegraphics[width=1\columnwidth]{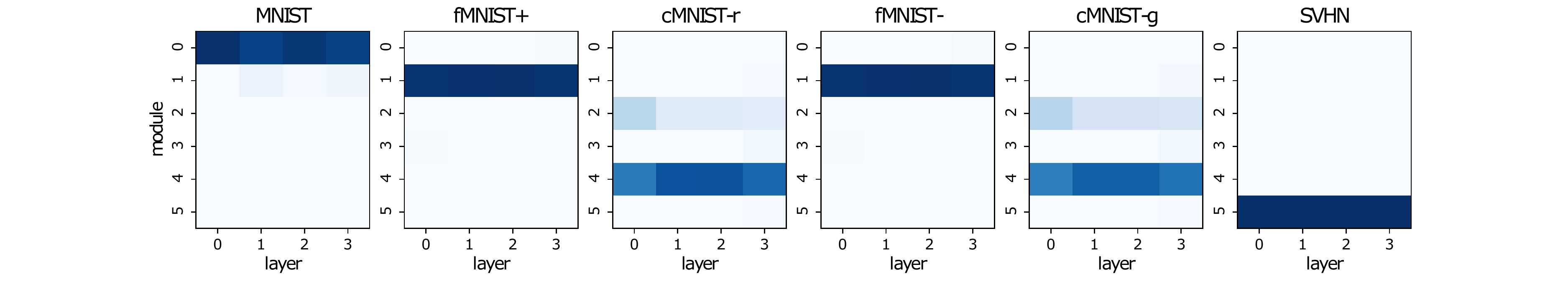}
    \caption{Average module selection per task on LMC3, which was constructed from plugging together independently trained LMC1 (first 3 tasks) and LMC2 (last 3 tasks).}%
\end{figure*}

\section{Compositional OOD generalization}\label{app:ood}
\begin{figure*}[!h]%
    \centering                                          
    \includegraphics[width=0.45\textwidth]{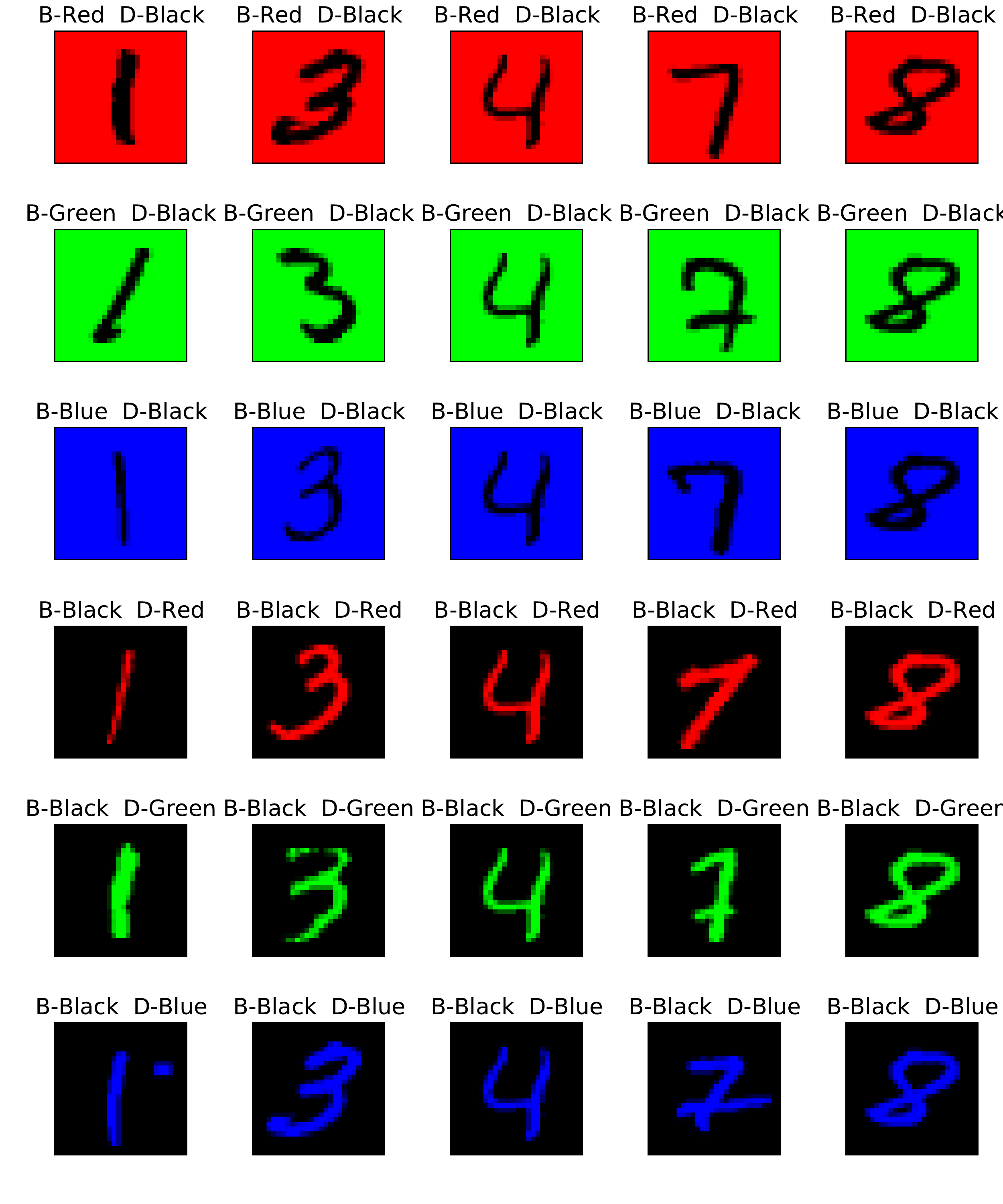}
    \caption{Examples of samples from the used colored-MNIST dataset.}%
    \label{fig:samples_cMNIST_ood}%
\end{figure*}

Some examples of the raw samples used for this experiment are presented in Figure~\ref{fig:samples_cMNIST_ood}. Model selection was performed assuming access to the test sets of the OOD tasks. As mentioned in \S~\ref{app:combining_independent_models}, this is sometimes an excepted strategy is settings in which test distribution is different from the training/validation distribution. Additionally, the input samples were normalized using statistics computed on training set of each tasks, including the OOD tasks (for which no training was performed).

In the main paper we presented a version of LMC with \textbf{omitted projection phase}. More precisely this means that the structural loss of the modules above was not propagated into the free modules on the lower layers (closer to the input) and that module addition was allowed during the whole training process.

\newpage
\subsection{Module selection}
\begin{figure*}[h!]%
    \centering             
    \includegraphics[width=0.9\textwidth]{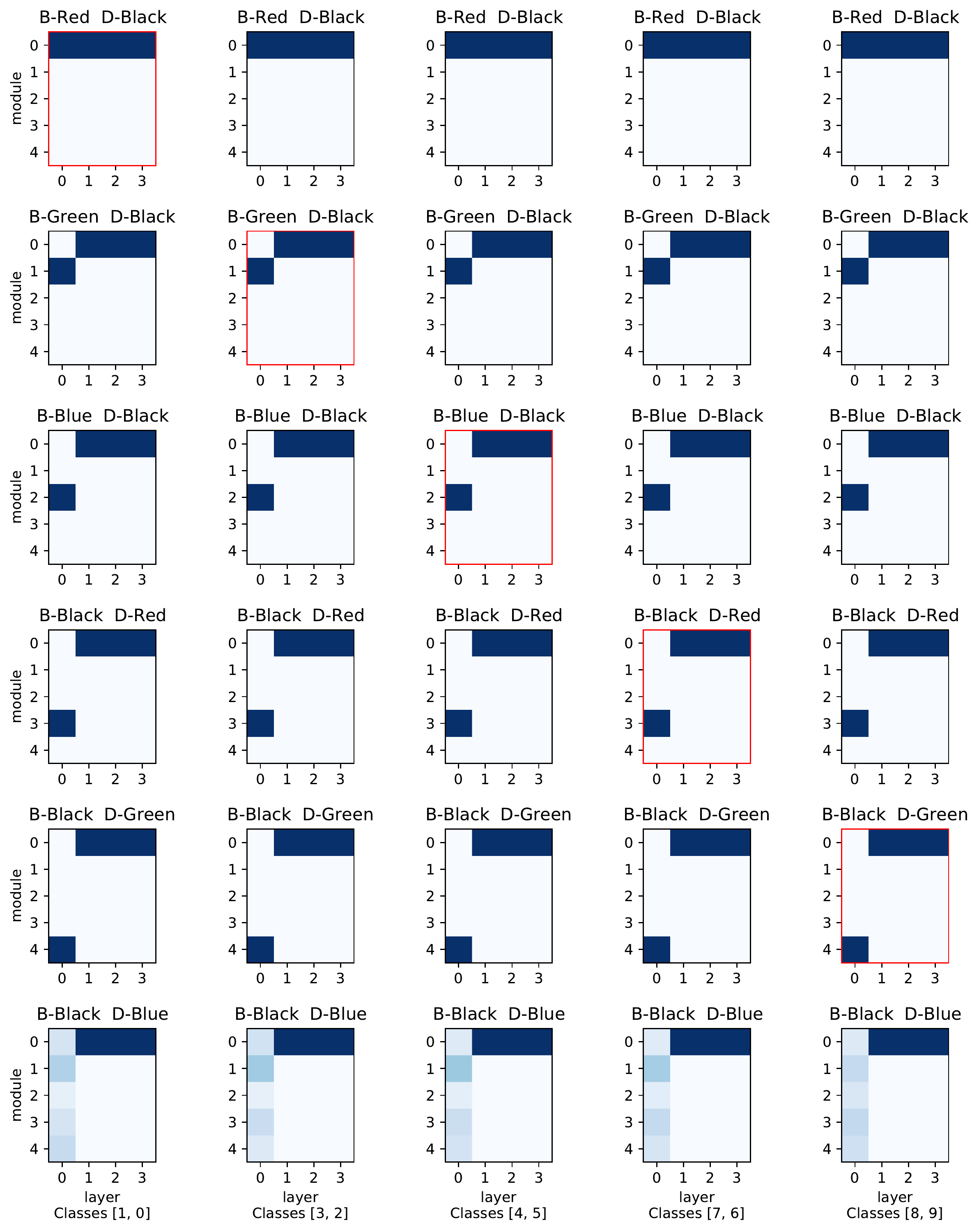}
    \caption{Average module selection per task for compositional OOD generalization setting presented in \S~\ref{sec:c_ood}. Each column represents a task (e.g. 0vs.1 classification), each row corresponds to a digit-background color combination. Each square represents a modular learner, x-axis is the layer, y-axis is the modules at each layer. The color intensities correspond to the average activation strength of the module when tested on the corresponding task (the darker the higher). \textbf{B} - stands for Background and \textbf{D} - stands for digit (e.g. ``B-Blue D-Black'' stands for blue background and black digit). The diagonal tasks (with red boarder) are seen during the continual training phase, after which the model is evaluated on all 5 combinations of seen digit-background colors for each task. The last row corresponds to the module selection for tasks with digit-background color combination which was not seen during the continual learning phase (black background+blue digit).}%
    \label{fig:module_selection_ood}%
\end{figure*}

In Figure~\ref{fig:module_selection_ood} we plot average module selection per task for the compositional OOD generalization setting.
\label{app:ood_gen}

\newpage
\section{Continual meta-learning}
\label{app:meta-cl}
In the continual meta-learning setting, a learner is exposed to tasks sampled from a sequence of environments. The goal of this setting is to construct a learner that quickly (i.e. within a few steps of gradient descent) learns tasks from new environments and relearns (or remembers) tasks from previously learned environments \citep{jerfel2019reconciling,caccia2020online,He2019TaskAC}. Methods applicable in this setting usually rely on gradient-based meta-learning strategies based on MAML~\cite{finn2017model}. The goal of MAML is to learn a good parameter initialization $\theta_0$ such that the learner can achieve low loss $\mathcal{L}_\tau$ on a randomly sampled task $\tau$ after a few steps of gradient descent. The objective of MAML can be formulated as \cite{nichol2018first}:
\begin{equation}
\theta_0 = arg\min_\theta \mathbb{E}_\tau(\mathcal{L}_\tau (U_\tau^k(\theta)) ),
\end{equation}
where where $U$ is the operator that performs $k$ inner-loop SGD updates starting with $\theta$ using data samples from task $\tau$. %
MAML has been extensively used in the few-shot learning scenario~\cite{wang2020generalizing}, where similar tasks are sampled from a stationary distribution. However, as pointed out by~\citet{jerfel2019reconciling}, it is unrealistic to assume the existence of a single set of meta-parameters that is close to all tasks in heterogeneous settings with outliers and non-stationary task distributions. In such setting, dissimilar tasks worsen generalization and non-stationarity causes catastrophic forgetting~\cite{kirkpatrick2017overcoming}. Instead of learning a single set of meta-parameters for all the tasks, \citet{jerfel2019reconciling} propose to learn a set of monolithic expert models, each representing a separate parameter initialization. The meta-learner then leverages the connection between gradient-based meta-learning and hierarchical Bayes~\cite{grant2018recasting} for selecting the most likely expert for the given task at hand. The probability of an expert to be selected is calculated based on the inner-loss of each expert as well as a non-parameteric Dirichlet process prior estimated in an online fashion. 

Here, we construct two environment sequences. In the first one, the learner is meta-trained on a sequence of Omniglot \cite{lake2015human}, MNIST \cite{lecun-mnisthandwrittendigit-2010} and fashion-MNIST \cite{xiao2017fashion} datasets --- the OMF sequence. We evaluate models in two settings. In the first setting, we report the average meta-test test accuracy over the environments seen so far. In the second setting, models are evaluated on environment  mixtures, i.e. each test task is composed of samples from the environments learned so far. A possible mixture task could contain classes 1 and 2 sampled from Omniglot, 2 and 3 from MNIST and class 5 sampled from the fashion-MNIST environment. The second sequence of environments is constructed from classes of the \textit{mini}Imagenet~\cite{ren2018meta}. 

We compare to the following baselines. \textbf{MAML}: non-modular MAML~\cite{finn2017model}. \textbf{MAML + ER}: MAML with experience replay --- we maintain a memory buffer of tasks sampled from previously seen environments, using a reservoir sampling procedure we retrieve a set of previously seen tasks which we then add to the task batches used for meta-training on new environments; \textbf{Meta-gating} tries to learn the relevance scores $\gamma^{(l)}$, used for gating, directly in the inner-loop. This is similar to soft-layer ordering used in~\citep{mendez2020lifelong}; 
\textbf{Experts}: an upper bound where separate monolithic model is trained for each environment and selected using the environment ID. We test LMCs with different levels of modular granularity: \textbf{LMCe} refers to a model that trains separate experts, but expert selection is performed using an  expert-level structural component; \textbf{LMCm} is the modular LMC with module selection performed at each layer (as described in \S~\ref{sec:lmc}). In our experiments we meta-learn the structural components: in the inner-loop we update functional parameters $\theta_m$ of all modules (also the fixed ones) whereas the structural parameters $\phi_m$ are kept constant, in the outer-loop both functional and structural parameters of the free modules are updated. \textit{Importantly, we do not apply the projection phase in the meta-learning experiments.} All experiments are performed in a 5-way 5-shot setting. 

For LMCm we used encoder-decoder architectures in the \textit{mini}Imagenet experiments (\S~\ref{app:imnet_experiments}), while for the OMF sequence the invertible structural component was used for the entire network. Note, that task inference is done automatically using the provided meta-test train (query) data, hence, there is no need to use multiple classification heads. For the OMF task sequence each module consists of a single convolutional layer with 64 3x3 filters (padding 1), batch-normalization, ReLU activation function and a max-pooling layer with the kernel-size of 2. This resembles the architecture introduced by \citet{finn2017model}. One epoch of meta-training consists of 100 meta-updates, each performed on a batch of 25 tasks (5-way, 5-shot). For the \textit{mini}ImageNet experiment, a single convolutional layer contains 32 filters. One epoch of meta-training performs 100 meta-updates, each on a batch of 4 tasks (5-way 5-shot regime). All learners contain 4 layers, non-modular (MAML, MAML+ER) learners do not expand. In the ``Experts'' baseline, each expert network corresponds to a 4-layered net with a single module per layer.

Model selection was performed using average meta-test validation accuracy over all environments. We randomly selected 10\% of the train datasets for validation purposes for MNIST and fMNIST datasets. For these datasets we did not evaluate the meta-generalization ability of the model, since the classes in meta-train and meta-test splits are the same. For the Omniglot dataset we selected 100 classes for validation purposes. We also flipped the background color of the Omniglot dataset to be black, which corresponds to the background colors of the MNIST and fMNIST datasets. For the \textit{mini}ImageNet dataset we split the 100 classes in 64-train/16-validation/20-test as in~\citep{ren2018meta}.

\subsection{Module addition continual meta-learning.}
\label{sec:module_addition_tricks}
In this section we explain some details about the adaptation of LMC's expansion strategy (\S\ref{sec:expansion_strategy}) for continual meta-learning.

The decision about module addition can be made per sample, per task, or per batch of tasks. In the per sample case, a new module is added whenever a sample is regarded as an outlier by all modules at a layer. In the per task case,  a new module is added whenever the average z-score for a task is larger than a threshold. In the per batch of tasks case, the average z-score is calculated over a batch of tasks. In our experiments we found that adding new modules on a per batch of tasks base yields the best efficacy in the continual meta-learning setting.

We do not freeze the modules at the environment switch, but create a checkpoint of each non-frozen (free)
module. Thus, modules are allowed to learn in the outer loop until the module addition is triggered. Whenever the module addition is triggered at a layer, the existing free module is dropped back to its state from the most recent checkpoint (i.e. its state at last environment switch).

\subsection{OMF results}

\begin{wrapfigure}[21]{t}{.40\textwidth}
    \begin{minipage}{\linewidth}
    \centering                              
    \includegraphics[width=\linewidth]{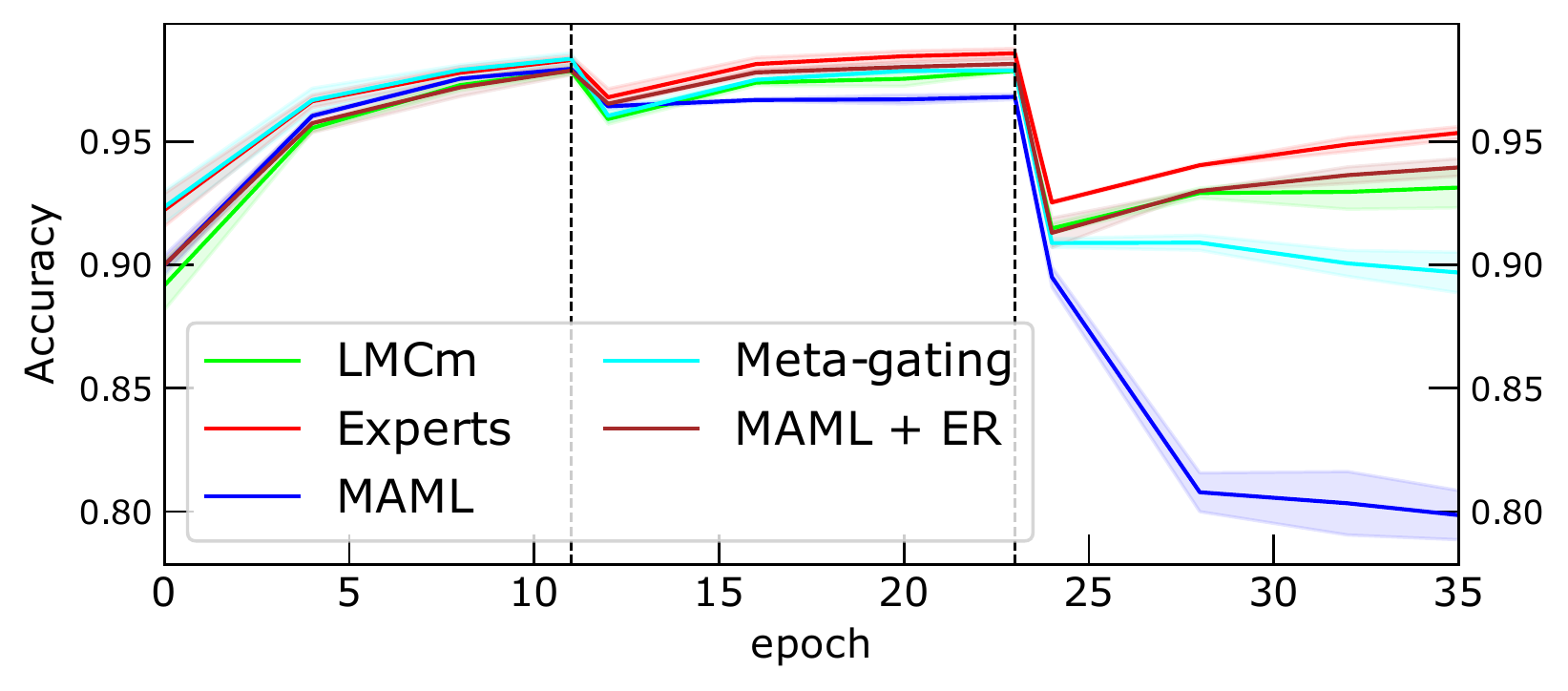}
    \subcaption{Average meta-test accuracy so far}
    \includegraphics[width=\linewidth]{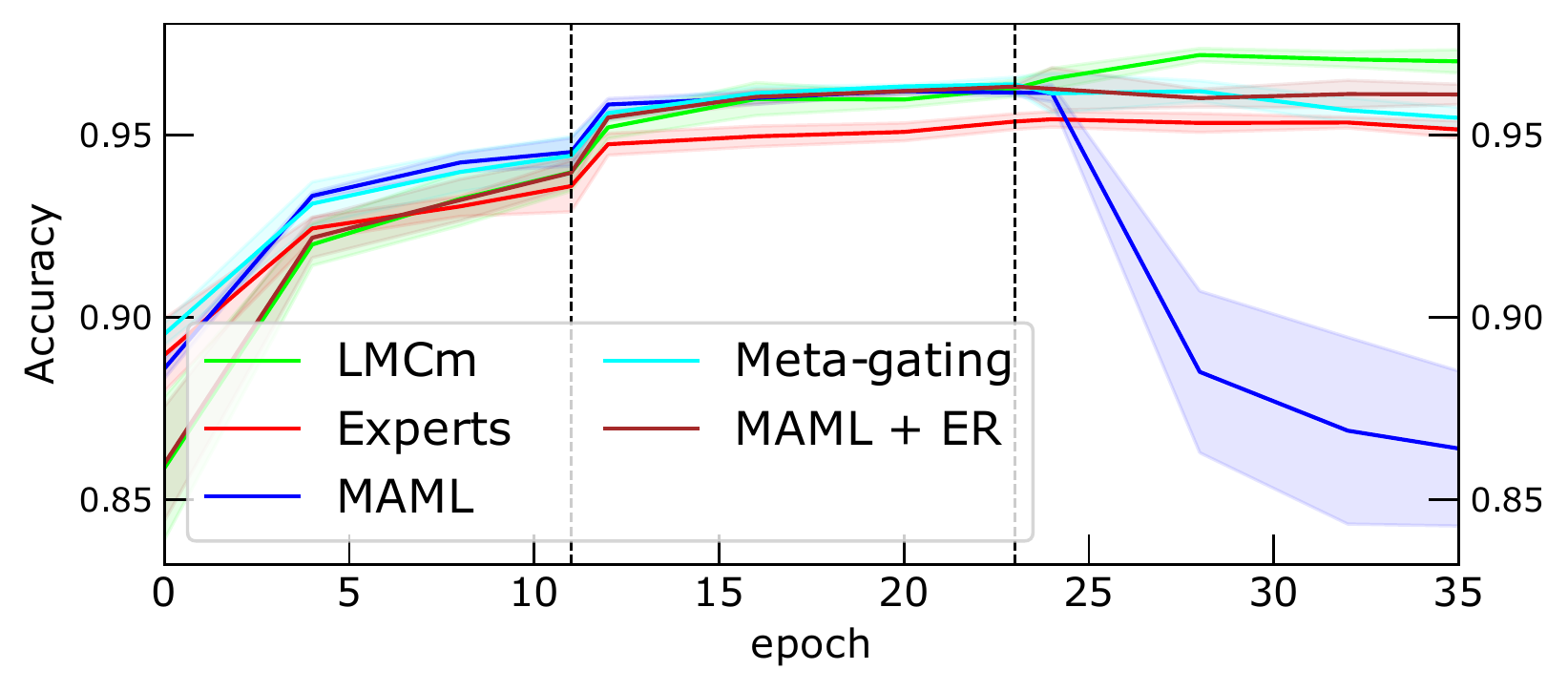}
    \subcaption{Mixtures of tasks meta-test accuracy}
\end{minipage}
\caption{Results on the OMF sequence. Vertical dashed lines mark the environment switches.}
\label{fig:acc_mixture_all}
\end{wrapfigure}

Results on OMF are shown in Figure \ref{fig:acc_mixture_all}. These results suggest that LMC can successfully avoid catastrophic forgetting achieving final average accuracy comparable to the experience replay (MAML + ER) baseline. This can be mainly attributed to the fact that learned modules at a layer are frozen every time a new module is added --- modules are not updated on new tasks if these tasks have triggered module addition and thus were recognized as outliers at a given layer.

Results with mixtures of tasks are in Figure~\ref{fig:acc_mixture_all}b. We find that LMC attains better performance than the other baselines. The performance improvement is due to LMC assigning modules on a per-sample basis in each forward pass, effectively guiding the inner- and outer-loop gradients of the meta-training procedure to the modules with higher activation on the current input. This contrasts the selection mechanism proposed by \cite{jerfel2019reconciling} for mixture-of-experts, where estimating the probability of each expert involves computation of the inner-loss which can be performed only on a per-task level. 

While replay slightly underperforms LMC on the mixture task, MAML is unable to reach high accuracy due to catastrophic forgetting. Regarding the experts baseline, each expert only specializes on one of the three environments, which results in low performance on the mixture-task. 

\begin{figure*}[!h]%
    \centering             
    \includegraphics[width=1\textwidth]{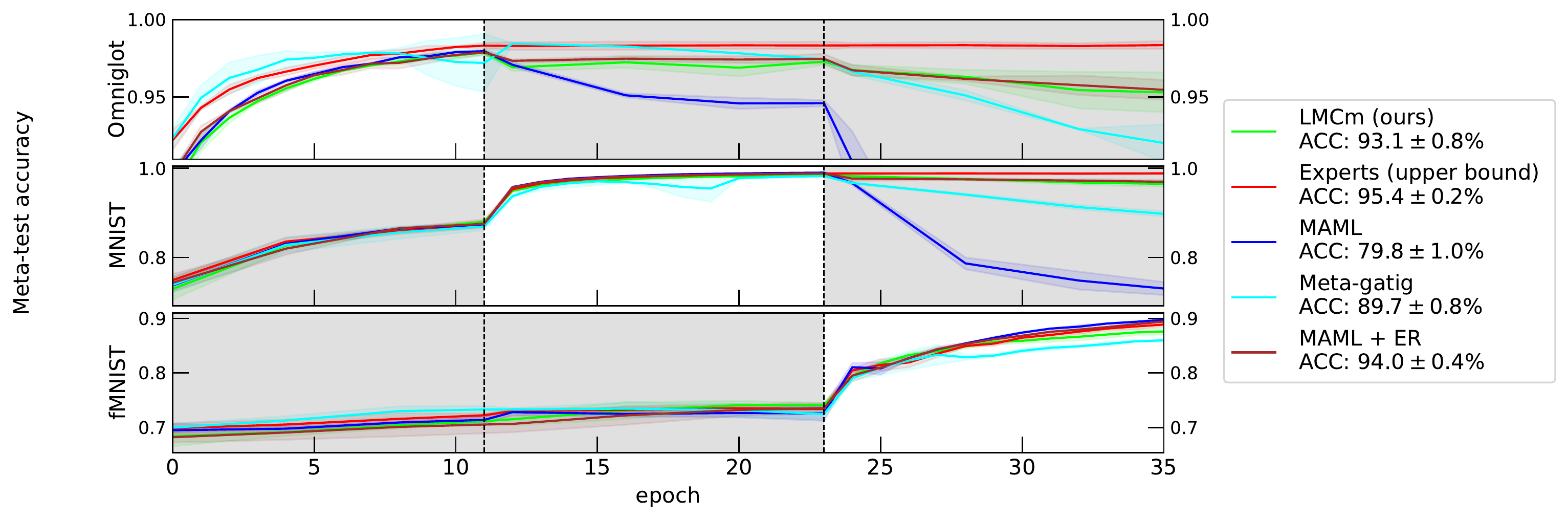} 
    \caption{Meta-test test accuracy on the OMF task sequence. Each row shows accuracy on a specific environment, the training time on each environment is highlighted with white background. We also report the average accuracy over all environments at the end of training in the legend (ACC). For this experiment the modular learner (LMCm) was trained using inevitable network as structural component (5-way 5-shot).}%
    \label{fig:omf_per_task}%
\end{figure*}

\begin{figure*}[!h]%
    \centering             
    \includegraphics[width=0.45\textwidth]{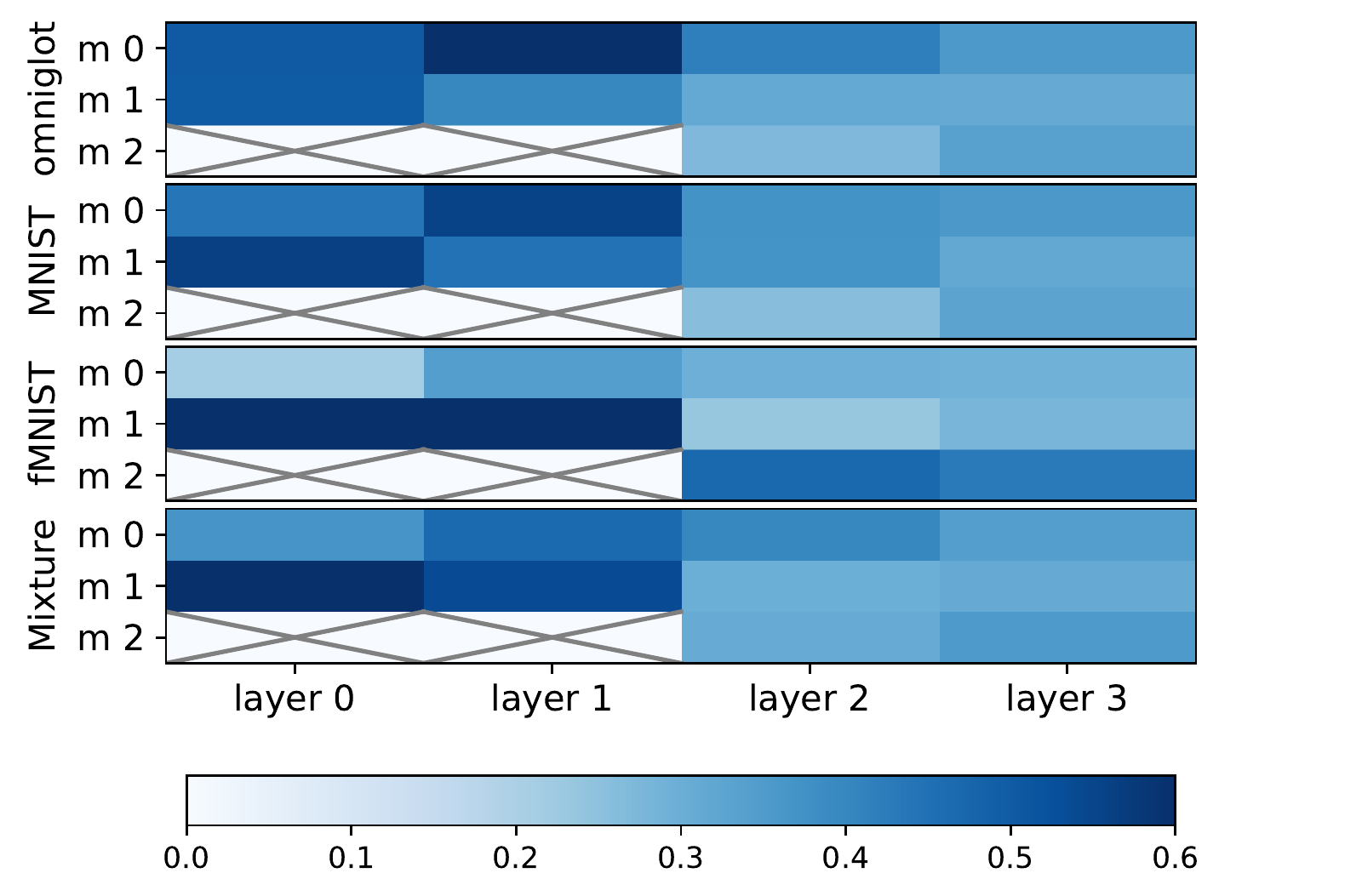}
    \caption{Average module selection of LMCm for a selected run on the OMF task sequence (after continual meta-CL training has been complete on the entire sequence).}
\label{fig:omf_module_selection}
\end{figure*}

In Figure~\ref{fig:omf_per_task} we show how the meta-test test accuracies evolve over the course of continual meta-training for each environment of the OMF sequence. Additionally, in Figure~\ref{fig:omf_module_selection} we plot the average module selection at meta-test test time on the OMF sequence after the entire sequence has been learned.

\subsection{Additional \textit{mini}Imagenet results.}
\label{app:imnet_experiments}
\begin{figure}[!t]%
    \begin{subfigure}[t]{0.45\textwidth}
        \includegraphics[width=1\linewidth]{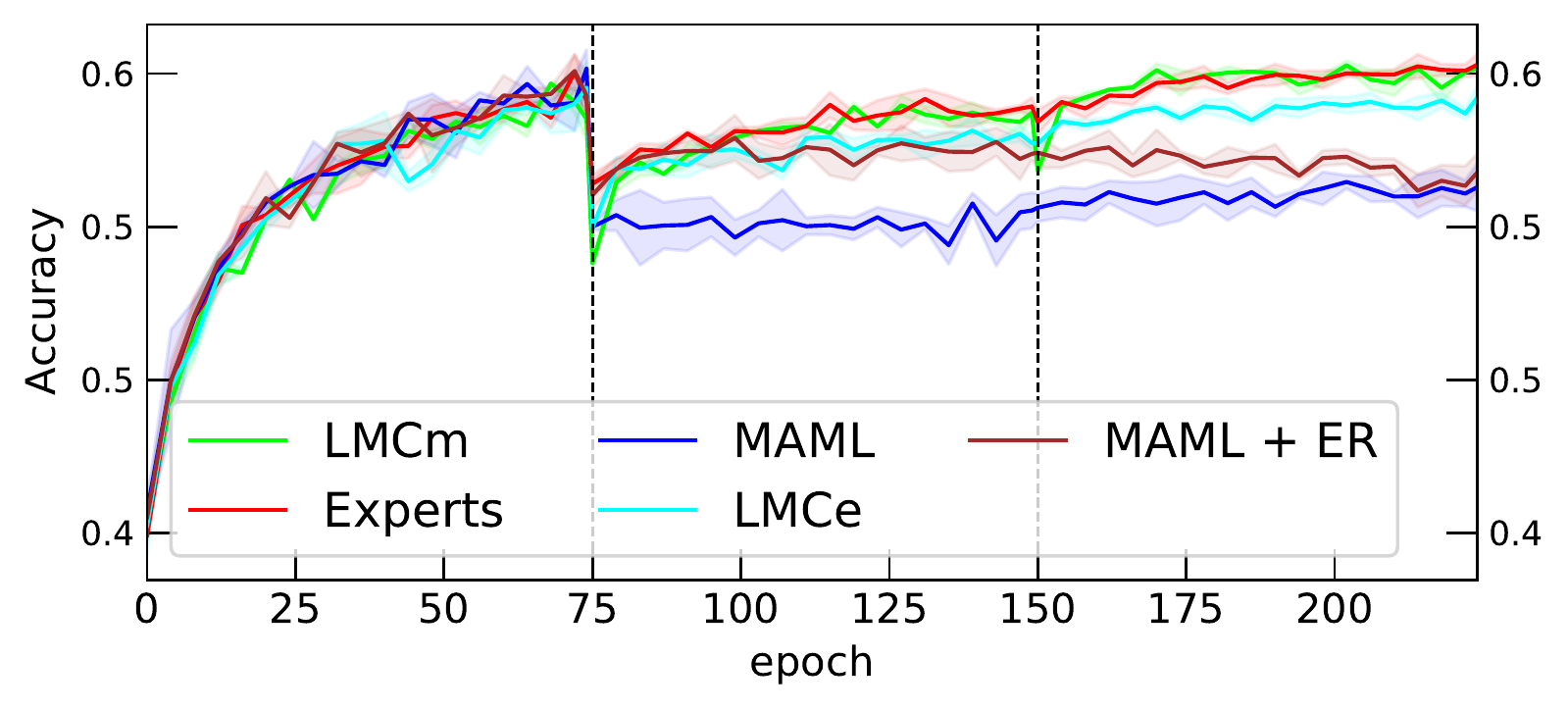}
        \caption{Average meta-test accuracy so far}
        \label{fig:ood-ewc}
    \end{subfigure}
    \begin{subfigure}[t]{0.45\textwidth}
        \includegraphics[width=1\linewidth]{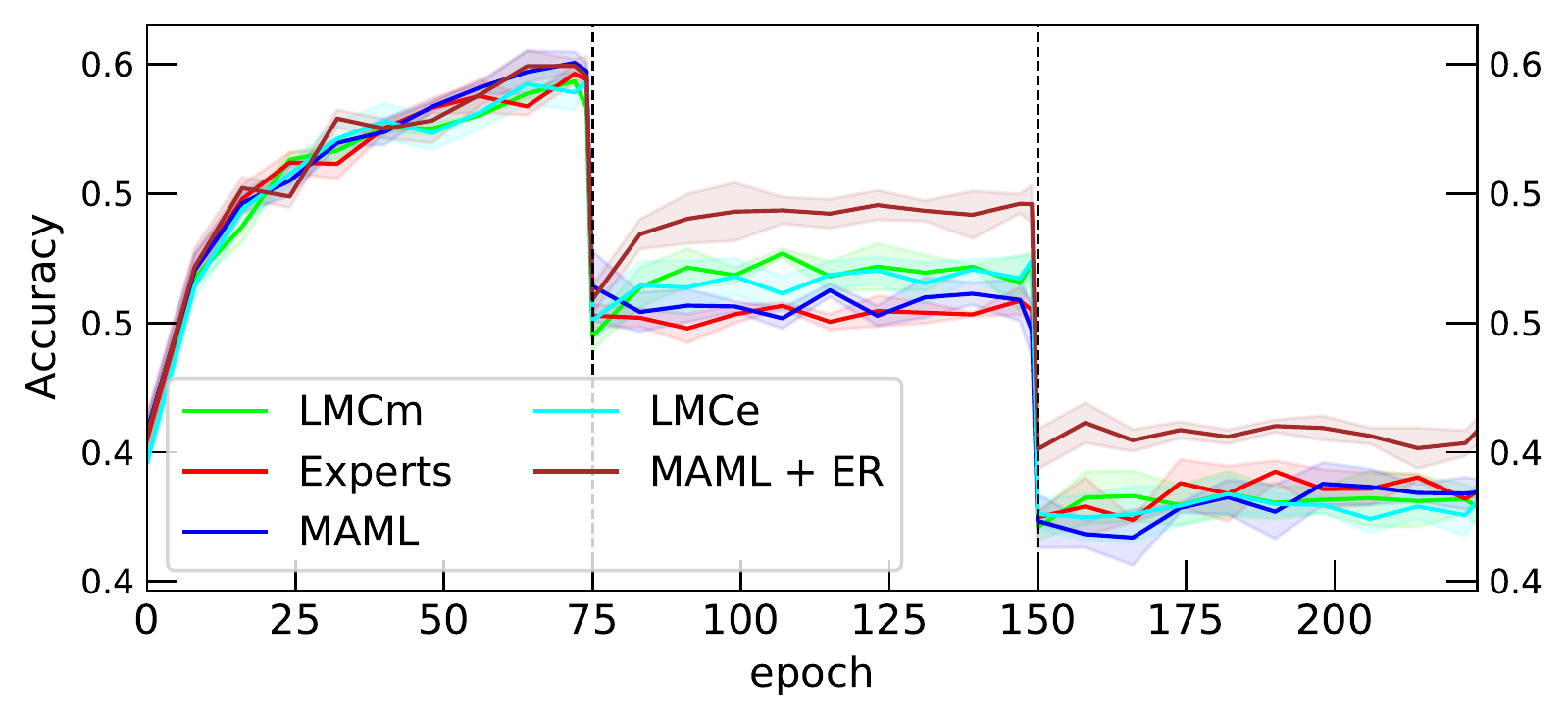}
        \caption{Average meta-test accuracy on the mixture tasks}
        \label{fig:ood-mntdp}
    \end{subfigure}
    \caption{Meta-test test results on the sequence of \textit{mini}Imagenet environments. Vertical lines mark environment switches.}%
    \label{fig:imnet_avverage_and_mixtures}
\end{figure}

\begin{figure*}[!h]%
    \centering           
    \includegraphics[width=1\textwidth]{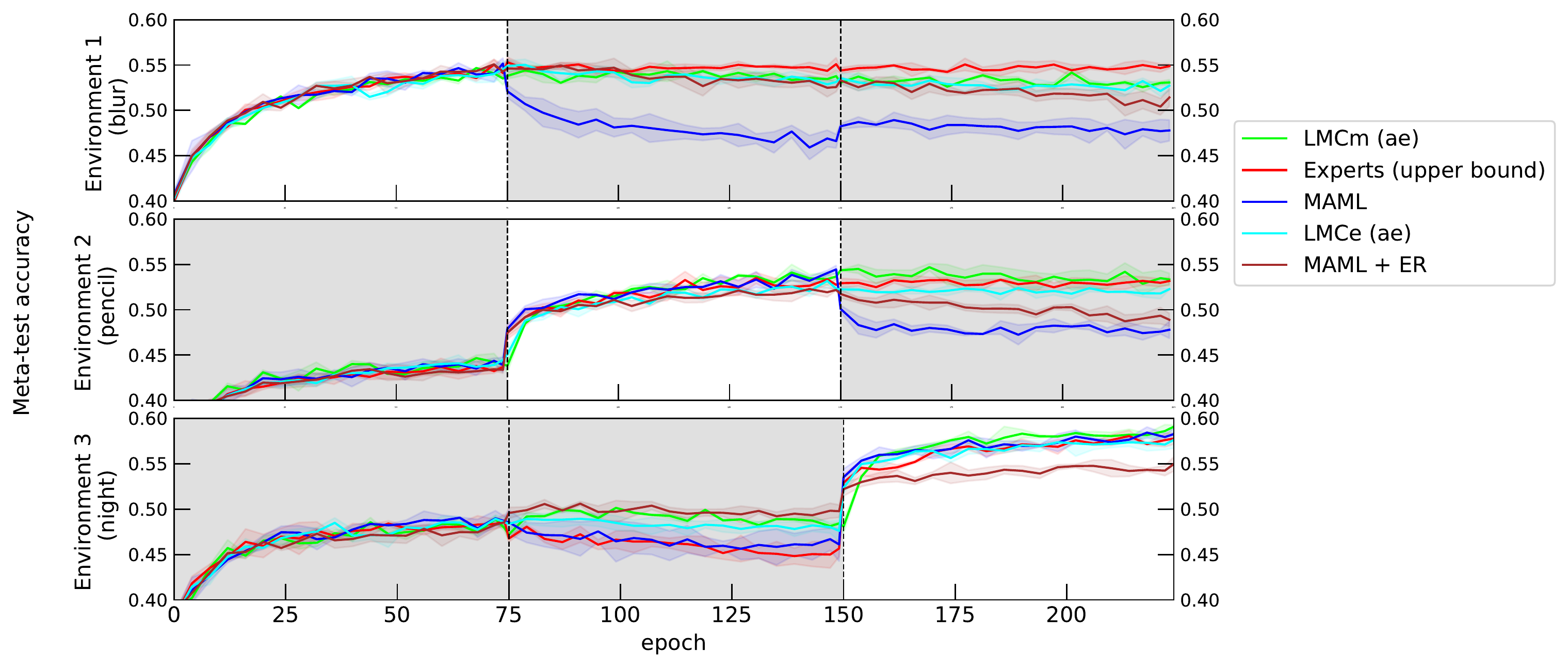}    
    \caption{Meta-test test accuracy on the evolving \textit{mini}ImageNet task sequence. Each row shows accuracy on a specific environment, the training time on each environment is highlighted with white background. For this experiment the modular learners (LMCm and LMCe) were trained using decoder as structural component tasked to reconstruct modules' inputs (5-way 5-shot).}%
    \label{fig:mini_imnet_per_task}%
\end{figure*}

In this section we present additional results on the evolving \textit{mini}Imagenet sequence where, similarly to~\citep{jerfel2019reconciling}, each environment is obtained through application of filters `blur', `pencil' and `night'. Figure~\ref{fig:mini_imnet_per_task} plots per environment accuracy of each method that results in the average accuracy over the environments seen so far depicted in Figure \ref{fig:imnet_avverage_and_mixtures} (a).

Additionally, we design a mixture-task baseline as follows: we select randomly 5 classes from the \textit{mini}ImageNet dataset, sample datapoints for these 5 classes from each of the environments seen so far (i.e. all three at the end of the sequence) to build tasks. Hence, a single mixture task for each class will contain samples from each environment seen so far.  The learner is then meta-tested on these tasks. This is different from the mixture task designed for the OMF sequence, where classes from different environments were mixed. We present the results on the mixture task in Figure~\ref{fig:imnet_avverage_and_mixtures}(b). We observe that MAML+ER baselines outperforms other learners in this evaluation setting. We hypothesize that poor performance of LMC in this setting is due to the batched modularity procedure (\S\ref{app:batched_module_selection}), where biasing module activation towards majority activation withing the batch of may harm performance, since each batch contains samples from all environments seen so far.

\section{Limitations}
\label{sec:limitations}     
We identify three main limitations of LMC. The first one is that its computational footprint is not constant with respect to the number of tasks. This is caused by the fact that each forward path through the modular learner requires a forward pass through each module in order to obtain the local relevance scores. Potentially, this can be overcome by assuming local stationarity and using the same modules for several consecutive batches of samples. On the other hand, this might not be a big problem at all: we think of each module as an independent entity that can be executed on it's own dedicated device, in which case at each layer each module's computations can be executed in parallel. The recent trend towards shifting from monolithic to modular architectures can accelerate the development of such hardware devices.

Another limitation of LMC is the lack of direct communication between modules at the same layer, which was shown to be important in some situations such as when different physical processes interact  \citep{goyal2019recurrent,goyal2021coordination}. While explicitly modeling such communication is an interesting direction for the future work, the consolidation of modules through weighed sum in Eq.~\ref{eq:x_l} can be though of as a form of implicit communication, i.e. a form of a shared workspace similar to the one presented in~\citep{goyal2021coordination}. It is however questionable whether cross-module communication is of any benefit in a standard supervised-learning settings.

Additionally, as discussed in \S~\ref{sec:longer_sequence}, module selection becomes challenging for LMC in presence of large number of candidate modules resulting in lower accuracy on long task sequences as compared to oracle based sleection strategy implemented in MNTDP~\cite{veniat2020efficient}.

Finally, in its working LMC relies on the local OOD detector as well as a generative model as it's structural component. Several recent works have shown that deep generative models often mistakenly assign high likelihood values to outlier points~\cite{nalisnick2018deep,hendrycks2018deep, wang2020further}. Hence, LMC's success depends on overcoming these issues through innovation in the fields of OOD detection and generative modeling.

\section{Broader societal impact}\label{app:societal_impact}
This work aimed at leveraging modularity and compositionality for continual learning (CL). The goal of CL is to design systems capable to retain knowledge and transfer knowledge across tasks. Such systems can \textbf{positively impact} society in the following ways: (i) models able to retain knowledge withing neural connections do not require storing raw samples in a replay buffer, yielding systems that are more compliant with data privacy standards. (ii) Positive transfer of knowledge across tasks can result in more resource efficient training. (iii) Building modular systems can further improve resource efficiency: e.g. as shown in the experiment in \S~\ref{sec:pnp} several modular systems can be combined in a third system without any retraining.

We do not identify any potential \textbf{negative societal impacts} of this work in particular beyond the potential negative societal impacts of artificial learning systems in general, which include the risk of decision bias, loss of certain jobs due to automation, risk of increased vulnerability to hacker attacks to name a few.

\end{document}